\documentclass{article} 
\usepackage{iclr2027_conference,times}

\usepackage{amsmath,amsfonts,bm}

\def\eqref#1{equation~\ref{#1}}

\def\1{\bm{1}}

\DeclareMathAlphabet{\mathsfit}{\encodingdefault}{\sfdefault}{m}{sl}
\SetMathAlphabet{\mathsfit}{bold}{\encodingdefault}{\sfdefault}{bx}{n}

\usepackage{hyperref}
\usepackage{url}
\usepackage{amssymb}
\usepackage{multirow}
\usepackage{tabularx}
\usepackage{colortbl}
\usepackage{graphicx}
\usepackage{wrapfig}
\usepackage{amsmath}
\usepackage{adjustbox}
\usepackage{array}
\usepackage{booktabs}
\usepackage{stfloats}
\usepackage{caption}

\title{Rethinking Fine-Tuning: Unlocking Hidden Capabilities in Vision-Language Models}

\author{
\begin{tabular}{@{}l@{}}
\textbf{Mingyuan Zhang$^{1}$, Yue Bai$^{1}$, Yifan Wang$^{1}$, Yiyang Huang$^{1}$, Yun Fu$^{1,2}$}
\end{tabular}\\[4pt]
$^{1}$Department of Electrical and Computer Engineering, Northeastern University\\
$^{2}$Khoury College of Computer Science, Northeastern University
}

\iclrfinalcopy 
\begin{document}

\maketitle
\lhead{}
\renewcommand{\headrulewidth}{0pt}

\begin{abstract}
Fine-tuning has become the dominant paradigm for adapting Vision-Language Models (VLMs), yet most approaches rely on explicit weight updates that introduce a fundamental trade-off. Full Fine-Tuning (FFT) may perturb pretrained representations due to cross-modal gradient interference, whereas Parameter-Efficient Fine-Tuning (PEFT) methods rely on additive modules, such as low-rank adapters, which may limit adaptation capacity.
In this paper, we rethink VLM adaptation from a structural selection framework that adapts VLMs without modifying backbone weights, and we propose \textbf{Mask Fine-Tuning (MFT)}. MFT learns masks that selectively route information through existing pretrained connections, dynamically uncovering subnetworks that better align pretrained representations with downstream objectives.
Extensive experiments show that MFT provides an effective structural alternative to both FFT and PEFT, consistently achieving superior performance across multiple vision-language benchmarks without adding knowledge or altering the deployment architecture. Moreover, our analysis with MFT provides new insights into how pretrained VLMs reorganize their internal representational pathways during adaptation. Codes: \emph{\href{https://github.com/Ming-K9/MFT-VLM}
{\textcolor{magenta}{https://github.com/Ming-K9/MFT-VLM}}}.
\end{abstract}

\section{Introduction}
\label{sec:intro}
\begin{wrapfigure}{r}{0.48\columnwidth}
    \centering
    \vspace{-10pt}
    \includegraphics[width=\linewidth]{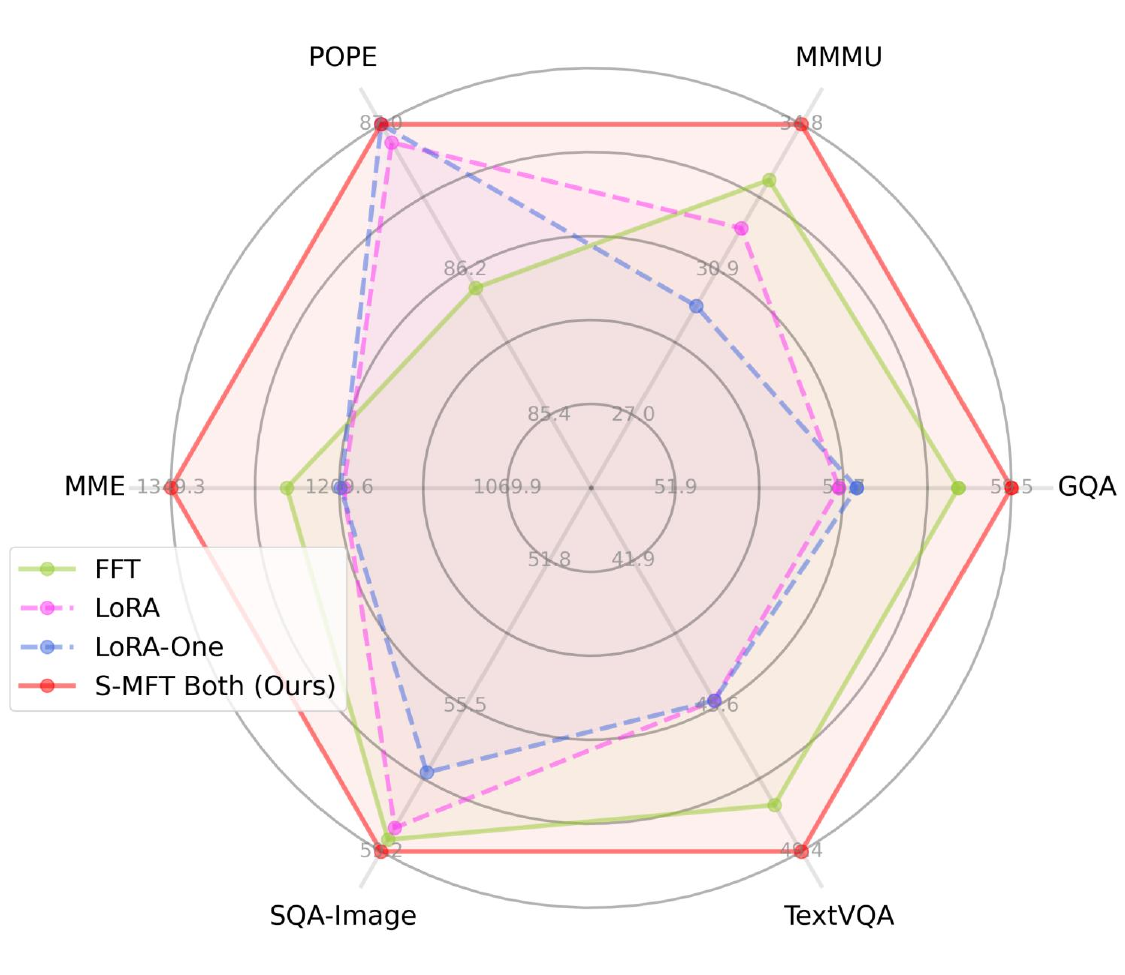}
    \vspace{-20pt}
    \caption{
    Comparison of fine-tuning methods on a VLM (Qwen2.5-0.5B language model).
    }
    \label{fig:intro_teaser}
    \vspace{-10pt}
\end{wrapfigure}
The advent of pre-trained Vision-Language Models (VLMs)~\citep{liu2023llava} represents a significant milestone in artificial intelligence~\citep{bordes2024introduction}.
By training on large-scale multimodal data, these models have demonstrated remarkable capabilities and set state-of-the-art standards across a spectrum of tasks, including visual question answering (VQA)~\citep{agrawal2016vqavisualquestionanswering}, image captioning~\citep{li2022blip}, and complex multimodal learning~\citep{li2025surveystateartlarge}. The VLM ecosystem is expanding at an unprecedented rate, with the proliferation of powerful open-weight language models~\citep{bai2023qwen,touvron2023llamaopenefficientfoundation,team2024gemma,gunasekar2023textbooks,zheng2023judging} built with fantastic vision backbones~\citep{radford2021learningtransferablevisualmodels,zhai2023sigmoidlosslanguageimage}.
However, despite their impressive zero-shot generalization, achieving peak performance in downstream applications, such as medical diagnostics~\citep{pham2025rarlimprovingmedicalvlm}, robotics~\citep{xu2025vlmadendtoendautonomousdriving}, or domain-specific document understanding~\citep{kim2022donut}, usually requires additional adaptation.

Traditionally, Full Fine-Tuning (FFT) has been the default paradigm~\citep{devlin-etal-2019-bert,ouyang2022traininglanguagemodelsfollow}. However, at model scale, FFT has become computationally and logistically prohibitive, requiring massive computational resources and storage for each downstream task~\citep{zhang2025parameterefficientfinetuningfoundationmodels}. To address this, Parameter-Efficient Fine-Tuning (PEFT)~\citep{zhang2025parameterefficientfinetuningfoundationmodels} methods, notably Low-Rank Adaptation (LoRA)~\citep{hu2021loralowrankadaptationlarge}, have been proposed. Although these additive PEFT paradigms balance effectiveness and efficiency, they introduce adaptation via additional low-rank modules, which may limit the capacity to model complex multimodal interactions.

In this paper, we rethink VLM adaptation not as an additive optimization problem, but as a structural selection problem over pretrained connections. While mask-based subnetwork discovery has been explored for narrow vision or language tasks~\citep{zhao-etal-2020-masking,RMT2023,fang2024maskllm}, its potential for adapting models from pre-trained checkpoints remains largely unexplored. We introduce Mask Fine-Tuning (MFT), which selectively reorganizes the overlapping representational pathways already encoded within the frozen VLM backbone. Rather than introducing external modules, MFT adapts the model by learning static masks that attenuate selected pre-trained connections. We conduct extensive experiments across diverse VLMs. The results show that MFT consistently and significantly surpasses strong baselines.
Our contributions are as follows:
\begin{itemize}
    \item We reconceptualize masking as a structural adaptation mechanism for VLMs, framing fine-tuning as a structural selection problem over pretrained connections.
    \item We propose Mask Fine-Tuning (MFT), a structural selection framework that adapts VLMs through trainable mask scores while keeping the pre-trained backbone weights frozen and adding no deployment-time parameters.
    \item Masks learned by MFT reveal structured sparsity patterns within VLMs, providing new insights into how pretrained models reorganize internal pathways during adaptation.
\end{itemize}

\section{Related Work}
\label{sec:related-work}
We contextualize Mask Fine-Tuning (MFT) within the broader landscape: 1) the evolution of fine-tuning methods for VLMs, 2) distinguishing MFT from model pruning, and 3) relevant mask-learning for fine-tuning methods.

\subsection{Vision-Language Models Fine-tuning}
Full Fine-Tuning (FFT) has been proven to be an effective downstream task adaptation mechanism of VLMs~\citep{oh2024calibratedrobustfinetuningvisionlanguage,zhai2024finetuning}. However, the prohibitive costs of FFT have become a problem as model size scales up~\citep{zhang2025parameterefficientfinetuningfoundationmodels}, leading to the rise of PEFT~\citep{zhang2025parameterefficientfinetuningfoundationmodels}. The vast majority of PEFT techniques, while diverse, share the assumption that adaptation requires introducing a weight update to the model. Adapter-Tuning~\citep{houlsby2019parameter} inserts small, bottleneck-shaped, and learnable neural modules (adapters) between the network layers. Prefix-Tuning~\citep{li2021prefixtuning} prepends a sequence of continuous and learnable vectors to the keys and values of the attention layers. Prompt-Tuning~\citep{lester-etal-2021-power} simplifies Prefix-Tuning by adding learnable embeddings only to the input layer. P-Tuning~\citep{liu-etal-2022-p} introduces a prompt encoder to more stably optimize continuous prompts. Low-Rank Adaptation (LoRA)~\citep{hu2021loralowrankadaptationlarge} is based on the hypothesis that the update matrix has a low intrinsic rank. It freezes the pre-trained weights and injects a trainable low-rank branch. This additive weight-update philosophy is shared by variants such as DoRA~\citep{liu2024doraweightdecomposedlowrankadaptation}, which decomposes the update into magnitude and direction components, and VeRA~\citep{kopiczko2024veravectorbasedrandommatrix}, which shares low-rank matrices across layers. Instead of adding new parameters or updating weights to achieve adaptation, MFT adapts the model by learning masks and keeps the pre-trained weights frozen.

\subsection{Pruning and Latent Subnetworks}
MFT shares conceptual connections with model pruning and latent subnetworks, which examine how over-parameterized models encode adaptable functional pathways. Neural network pruning~\citep{liu2018rethinking,molchanov2017pruningconvolutionalneuralnetworks,wang2022recentadvancesneuralnetwork} has been widely adopted across vision~\citep{fang2024isomorphic}, language~\citep{liu2025pruning}, and multimodal learning~\citep{liu2025meteormultiencodercollaborativetoken} for compression~\citep{han2016deepcompressioncompressingdeep,iandola2016squeezenetalexnetlevelaccuracy50x} and acceleration~\citep{han2016eieefficientinferenceengine,wang2021neural,wang2023trainability,NEURIPS2023_44956951,NEURIPS2023_35c1d69d}. In parallel, the Lottery Ticket Hypothesis (LTH)~\citep{frankle2019lotterytickethypothesisfinding} revealed that dense networks can contain sparse subnetworks capable of matching dense performance. The Dual Lottery Ticket Hypothesis (DLTH~\citep{bai2022dual}) further showed that targeted regularization can transform even randomly selected subnetworks into trainable ones. Similar findings in pre-trained models such as BERT~\citep{liu2022learningwinlotterytickets} indicate that transferable subnetworks exist and can be selectively activated for new tasks in large models, consistent with MFT's view of reusing existing structures in foundation models with frozen weights. 
While both pruning and MFT involve identifying meaningful connections, their fundamental objectives and underlying mechanisms diverge sharply. Pruning and traditional LTH searches focus strictly on model compression, aiming to reduce model size and inference latency via permanent sparsity~\citep{frantar-sparsegpt,sun2024simpleeffectivepruningapproach}, often accompanied by performance degradation. In contrast, MFT treats masking not as a compression-oriented tool, but as an \textit{end-to-end structural selection mechanism} designed to facilitate cross-modal interactions. Rather than reducing parameters, MFT is designed for knowledge elicitation, removing redundancy and improving performance.

\subsection{Subnetwork and Mask Learning Adaptation}
MFT belongs to a new and emerging class of methods focusing on subnetwork identification, in which adaptation is achieved solely by learning a mask over a completely frozen backbone. This paradigm reframes model adaptation as a structural selection problem rather than parameter optimization, aligning with a broader trend of identifying optimal subnetworks within pre-trained models~\citep{fang2024maskllm,hou2025instructionfollowingpruninglargelanguage,liu2025proxsparse,tao-etal-2023-structured}. 
Early seminal works, such as~\citep{zhao-etal-2020-masking}, show that discrete binary masking of pre-trained BERT weights can achieve performance comparable to FFT. Regularized Mask Tuning (RMT)~\citep{RMT2023} extends this idea to the vision encoder in CLIP~\citep{radford2021learningtransferablevisualmodels}. 
While effective at identifying isolated subnetworks for narrow, discriminative tasks, these methods are architecturally misaligned with the complex, cross-modal generative instruction-following required by modern VLMs. Furthermore, recent methods~\citep{zhang2025boostinglargelanguagemodels} treat masking as a strictly post-hoc enhancement step. MFT distinguishes itself as an end-to-end, general-purpose instruction-tuning strategy that operates directly on base pretrained VLMs.

\section{Mask Fine-Tuning (MFT)}
\label{sec:method}
\begin{figure}[t]
    \centering
    \includegraphics[width=0.95\columnwidth]{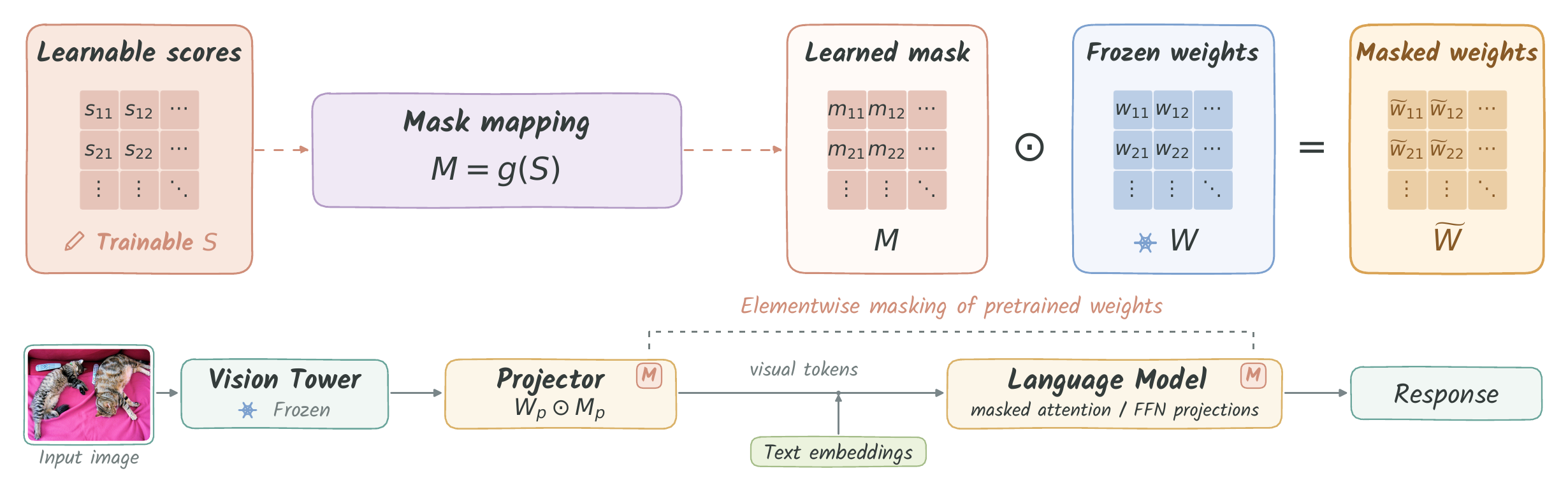}
    \vspace{-4pt}
    \caption{
    Overview of \textbf{Mask Fine-Tuning (MFT)}. All pretrained parameters remain frozen, while learnable scores $S$ are optimized for the projector and selected linear layers of the language backbone. The scores are mapped to masks $M=g(S)$, which modulate the pretrained weights $W$ through elementwise multiplication, yielding the effective weights $\widetilde{W}=W\odot M$.
    }
    \label{fig:method-arch}
    \vspace{-10pt}
\end{figure}

We adopt an MFT approach for VLMs, specifically learning masks for the language backbone and projector components, shown in Fig.~\ref{fig:method-arch}.

\subsection{Mask Parameterization}
For each pre-trained weight matrix $W \in \mathbb{R}^{m \times n}$ in the language model and projector, we introduce a learnable score matrix of the same size $S \in \mathbb{R}^{m \times n}$ to help generate the mask $M \in \mathbb{R}^{m \times n}$. The effective weight is computed as:
\begin{equation}
W' = W \odot M,
\end{equation}
where $\odot$ denotes element-wise multiplication, $W$ is frozen during training.

\subsection{Learnable Mask}
\noindent\textbf{Binary (Hard) Mask.}\label{h-mft}
For a target sparsity level $k\in[0,1]$, set $s=\lfloor kmn\rfloor$. Let
$\mathcal I_s(S)\subseteq\{1,\ldots,m\}\times\{1,\ldots,n\}$
be the set of exactly $s$ indices having the smallest values of $|S_{ij}|$, using a fixed lexicographic rule to break ties. We define
\begin{equation}
M_{ij}=
\begin{cases}
0,&(i,j)\in\mathcal I_s(S),\\
1,&(i,j)\notin\mathcal I_s(S).
\end{cases}
\end{equation}
This definition masks exactly $s$ entries and retains exactly $mn-s$. The trainable score magnitude provides a task-aware ranking signal that directly determines the selected connections.

To optimize the exact top-$(1-k)$ binary selection, we employ the Straight-Through Estimator (STE)~\citep{bengio2013estimatingpropagatinggradientsstochastic} for gradient $\mathcal{L}$ computation, which enables gradients to flow directly through the discrete masking operation by approximating the Jacobian as an identity mapping:
\begin{equation}
\frac{\partial M}{\partial S} \approx I,
\end{equation}
where $I$ denotes the identity. While the forward pass enforces strict binary values via top-$k$ selection, the backward pass treats the operation as identity, enabling the scores $S$ to receive meaningful gradient signals. By the chain rule:
\begin{equation}
\frac{\partial \mathcal{L}}{\partial S} = \frac{\partial \mathcal{L}}{\partial M} \cdot \frac{\partial M}{\partial S} \approx \frac{\partial \mathcal{L}}{\partial M} \cdot I = \frac{\partial \mathcal{L}}{\partial M}.
\end{equation}
This approximation provides effective gradients for optimization, guiding which elements to mask.

\noindent\textbf{Continuous (Soft) Mask.}\label{sec:s-mft}
Binary masking provides exact sparsity and clear interpretability, while S-MFT complements it with a differentiable formulation that automatically calibrates masks using visual and linguistic signals. Based on a temperature-scaled sigmoid function, S-MFT smoothly learns task-adaptive structural selection patterns:
\begin{equation}
M_{ij} = \sigma\left(\frac{S_{ij}}{T}\right) = \frac{1}{1 + \exp(-S_{ij}/T)},
\end{equation}
where $T > 0$ is a static temperature parameter that controls the sharpness of the function. The sigmoid parameterization places every effective weight on the line segment between zero and its pre-trained value, preserving the pre-trained sign and preventing uncontrolled amplification. This structure-selection geometry supplies a focused inductive bias while retaining continuous, task-adaptive control over every selected connection.

\begin{table*}[t]
    \caption{
    Main experimental result comparison with the time cost of our Soft Mask Fine-Tuning (S-MFT) with Full Fine-Tuning (FFT) and representative Parameter-Efficient Fine-Tuning (PEFT) baselines across four language backbones. We repeat each experiment five times with different random seeds and report the mean and standard deviation. \textcolor[RGB]{235,85,75}{Red numbers} denote the best results, and \textcolor[RGB]{75,115,210}{blue numbers} denote the second-best. All results reported here are the performance after a full training epoch. ``Epochs'' and ``Time'' are reported as the efficiency metric, indicating when a method converges and achieves stable, optimal validation performance.
    }
    \vspace{-2mm}
    \centering
    \setlength{\tabcolsep}{8pt}
    \adjustbox{width=\textwidth}{
    \begin{tabular}{llccclcccc} 
        \toprule
        \multicolumn{2}{c}{\multirow{2}{*}{\textbf{Method}}} 
        & \multicolumn{6}{c}{\textbf{Evaluation Tasks}} 
        & \multirow{2}{*}{\textbf{Epoch} {↓}}
        & \multirow{2}{*}{\textbf{Time (h)} {↓}} \\ 
        \cmidrule(lr){3-8}
        \multicolumn{2}{l}{} 
        & \multicolumn{1}{c}{\textbf{GQA} {↑}} 
        & \multicolumn{1}{c}{\textbf{MMMU} {↑}} 
        & \multicolumn{1}{c}{\textbf{POPE} {↑}} 
        & \multicolumn{1}{c}{\textbf{MME} {↑}} 
        & \multicolumn{1}{c}{\textbf{SQA-Image} {↑}} 
        & \multicolumn{1}{c}{\textbf{TextVQA} {↑}}
        & 
        & \\ 
        \midrule
        \multirow{10}{*}{\rotatebox[origin=c]{90}{\textbf{Qwen2.5-0.5B}}}
        & Zero-Shot
        & 5.7 
        & 22.6 
        & 49.9 
        & \multicolumn{1}{c}{93.7} 
        & 3.2 
        & 9.0 
        & - 
        & - \\
        & FFT
        & 58.3$_{\pm0.28}$ 
        & 33.3$_{\pm0.37}$ 
        & 86.1$_{\pm0.46}$ 
        & 1253.0$_{\pm10.2}$ 
        & 58.9$_{\pm0.25}$ 
        & 48.2$_{\pm0.23}$ 
        & 0.90 
        & 4.91 \\
        & LoRA
        & 55.6$_{\pm0.28}$ 
        & 32.0$_{\pm0.35}$ 
        & 86.9$_{\pm0.45}$ 
        & 1206.0$_{\pm10.1}$ 
        & 58.6$_{\pm0.22}$ 
        & 45.5$_{\pm0.24}$ 
        & 0.65 
        & 3.14 \\
        & DoRA
        & 57.1$_{\pm0.24}$ 
        & 32.8$_{\pm0.31}$ 
        & \textcolor[RGB]{75,115,210}{\textbf{87.1}}$_{\pm0.42}$ 
        & 1228.5$_{\pm9.4}$ 
        & 58.8$_{\pm0.21}$ 
        & 46.8$_{\pm0.28}$ 
        & 0.63 
        & 3.42 \\
        & QLoRA
        & 53.8$_{\pm0.36}$ 
        & 30.1$_{\pm0.41}$ 
        & 85.2$_{\pm0.47}$ 
        & 1178.4$_{\pm10.2}$ 
        & 56.9$_{\pm0.27}$ 
        & 46.7$_{\pm0.39}$ 
        & 0.88 
        & 5.97 \\
        & Parallel Adapter
        & 56.0$_{\pm0.25}$ 
        & 30.8$_{\pm0.32}$ 
        & 86.5$_{\pm0.49}$ 
        & 1215.4$_{\pm9.5}$ 
        & 57.6$_{\pm0.20}$ 
        & 45.8$_{\pm0.23}$ 
        & 0.72 
        & 3.95 \\
        & P-Tuning v2
        & 52.8$_{\pm0.21}$ 
        & 28.5$_{\pm0.34}$ 
        & 84.1$_{\pm0.46}$ 
        & 1162.3$_{\pm10.9}$ 
        & 54.2$_{\pm0.29}$ 
        & 42.4$_{\pm0.39}$ 
        & 0.85 
        & 3.58 \\
        & \textbf{S-MFT Attn(Ours)} 
        & 58.0$_{\pm0.27}$ 
        & 33.3$_{\pm0.31}$ 
        & 86.7$_{\pm0.47}$ 
        & 1243.0$_{\pm8.2}$ 
        & \textcolor[RGB]{75,115,210}{\textbf{59.7}}$_{\pm0.20}$ 
        & 47.9$_{\pm0.37}$ 
        & \textcolor[RGB]{235,85,75}{\textbf{0.61}} 
        & \textcolor[RGB]{235,85,75}{\textbf{2.92}}\\
        & \textbf{S-MFT MLP (Ours)} 
        & \textcolor[RGB]{75,115,210}{\textbf{59.0}}$_{\pm0.21}$ 
        & \textcolor[RGB]{75,115,210}{\textbf{33.4}}$_{\pm0.32}$ 
        & \textcolor[RGB]{235,85,75}{\textbf{87.3}}$_{\pm0.45}$ 
        & \textcolor[RGB]{75,115,210}{\textbf{1306.0}}$_{\pm8.4}$ 
        & \textcolor[RGB]{235,85,75}{\textbf{60.0}}$_{\pm0.17}$ 
        & \textcolor[RGB]{75,115,210}{\textbf{48.9}}$_{\pm0.39}$ 
        & \textcolor[RGB]{75,115,210}{\textbf{0.62}} 
        & \textcolor[RGB]{75,115,210}{\textbf{3.13}} \\
        & \textbf{S-MFT Both(Ours)} 
        & \textcolor[RGB]{235,85,75}{\textbf{59.5}}$_{\pm0.22}$ 
        & \textcolor[RGB]{235,85,75}{\textbf{34.8}}$_{\pm0.34}$ 
        & 87.0$_{\pm0.48}$ 
        & \textcolor[RGB]{235,85,75}{\textbf{1349.3}}$_{\pm8.6}$ 
        & 59.2$_{\pm0.19}$ 
        & \textcolor[RGB]{235,85,75}{\textbf{49.4}}$_{\pm0.38}$ 
        & 0.67 
        & 3.81 \\
        \midrule
        \multirow{10}{*}{\rotatebox[origin=c]{90}{\textbf{TinyLlama-1.1B}}}
        & Zero-Shot
        & 1.0 
        & 26.1 
        & 49.8 
        & \multicolumn{1}{c}{211.8} 
        & 4.0 
        & 6.8 
        & - 
        & - \\
        & FFT
        & 57.3$_{\pm0.33}$ 
        & 30.4$_{\pm0.34}$ 
        & 85.8$_{\pm0.44}$ 
        & 1180.9$_{\pm9.2}$ 
        & 60.3$_{\pm0.27}$ 
        & 47.9$_{\pm0.23}$ 
        & 0.93 
        & 5.76 \\
        & LoRA
        & 57.5$_{\pm0.41}$ 
        & 30.4$_{\pm0.22}$ 
        & 86.9$_{\pm0.47}$ 
        & 1167.6$_{\pm9.5}$ 
        & 58.2$_{\pm0.19}$ 
        & 47.9$_{\pm0.27}$ 
        & 0.77 
        & 4.53 \\
        & DoRA
        & 57.4$_{\pm0.31}$ 
        & \textcolor[RGB]{75,115,210}{\textbf{30.9}}$_{\pm0.25}$ 
        & 87.1$_{\pm0.43}$ 
        & 1175.2$_{\pm9.8}$ 
        & 59.8$_{\pm0.18}$ 
        & 48.1$_{\pm0.26}$ 
        & 0.72 
        & 4.81 \\
        & QLoRA
        & 55.9$_{\pm0.35}$ 
        & 26.2$_{\pm0.35}$ 
        & 81.5$_{\pm0.48}$ 
        & 1142.8$_{\pm11.5}$ 
        & 56.5$_{\pm0.22}$ 
        & 46.2$_{\pm0.38}$ 
        & 0.62 
        & 5.03 \\
        & Parallel Adapter
        & 56.8$_{\pm0.30}$ 
        & 29.8$_{\pm0.31}$ 
        & 86.4$_{\pm0.48}$ 
        & 1165.5$_{\pm8.9}$ 
        & 57.5$_{\pm0.21}$ 
        & 46.8$_{\pm0.24}$ 
        & 0.81 
        & 4.62\\
        & P-Tuning v2
        & 53.4$_{\pm0.27}$ 
        & 25.1$_{\pm0.28}$ 
        & 82.5$_{\pm0.44}$ 
        & 1105.0$_{\pm11.8}$ 
        & 52.7$_{\pm0.14}$ 
        & 44.4$_{\pm0.22}$ 
        & 0.88 
        & 4.25 \\
        & \textbf{S-MFT Attn(Ours)} 
        & 57.1$_{\pm0.20}$ 
        & 30.7$_{\pm0.29}$ 
        & 87.6$_{\pm0.47}$ 
        & 1170.9$_{\pm7.4}$ 
        & 60.4$_{\pm0.15}$ 
        & \textcolor[RGB]{235,85,75}{\textbf{49.1}}$_{\pm0.37}$ 
        & \textcolor[RGB]{75,115,210}{\textbf{0.61}} 
        & \textcolor[RGB]{235,85,75}{\textbf{2.98}} \\
        & \textbf{S-MFT MLP (Ours)} 
        & \textcolor[RGB]{75,115,210}{\textbf{57.9}}$_{\pm0.30}$ 
        & \textcolor[RGB]{75,115,210}{\textbf{30.9}}$_{\pm0.20}$ 
        & \textcolor[RGB]{235,85,75}{\textbf{88.0}}$_{\pm0.48}$ 
        & \textcolor[RGB]{235,85,75}{\textbf{1237.9}}$_{\pm8.2}$ 
        & \textcolor[RGB]{235,85,75}{\textbf{62.6}}$_{\pm0.16}$ 
        & 47.4$_{\pm0.35}$ 
        & \textcolor[RGB]{235,85,75}{\textbf{0.60}} 
        & \textcolor[RGB]{75,115,210}{\textbf{3.65}} \\
        & \textbf{S-MFT Both(Ours)} 
        & \textcolor[RGB]{235,85,75}{\textbf{58.1}}$_{\pm0.22}$ 
        & \textcolor[RGB]{235,85,75}{\textbf{32.0}}$_{\pm0.32}$ 
        & \textcolor[RGB]{75,115,210}{\textbf{87.7}}$_{\pm0.47}$ 
        & \textcolor[RGB]{75,115,210}{\textbf{1218.0}}$_{\pm7.7}$ 
        & \textcolor[RGB]{75,115,210}{\textbf{61.0}}$_{\pm0.19}$ 
        & \textcolor[RGB]{75,115,210}{\textbf{48.7}}$_{\pm0.39}$ 
        & 0.74 
        & 4.32 \\
        \midrule
        \multirow{10}{*}{\rotatebox[origin=c]{90}{\textbf{Gemma-2B}}}
        & Zero-Shot
        & 2.6 
        & 26.4 
        & 56.6 
        & \multicolumn{1}{c}{586.2} 
        & 2.8 
        & 5.6 
        & - 
        & - \\
        & FFT
        & 60.9$_{\pm0.37}$ 
        & \textcolor[RGB]{75,115,210}{\textbf{31.3}}$_{\pm0.35}$ 
        & 86.4$_{\pm0.44}$ 
        & 1339.0$_{\pm10.8}$ 
        & 60.5$_{\pm0.27}$ 
        & 50.7$_{\pm0.26}$ 
        & 0.88 
        & 13.98 \\
        & LoRA
        & 58.3$_{\pm0.32}$ 
        & 29.6$_{\pm0.30}$ 
        & 86.2$_{\pm0.46}$ 
        & 1178.0$_{\pm9.7}$ 
        & 57.4$_{\pm0.20}$ 
        & 48.0$_{\pm0.23}$ 
        & 0.64 
        & 9.43 \\
        & DoRA
        & 59.6$_{\pm0.28}$ 
        & 30.8$_{\pm0.32}$ 
        & 86.8$_{\pm0.41}$ 
        & 1250.4$_{\pm9.5}$ 
        & 59.1$_{\pm0.22}$ 
        & 49.8$_{\pm0.24}$ 
        & 0.61 
        & 9.85 \\
        & QLoRA
        & 56.7$_{\pm0.36}$ 
        & 29.9$_{\pm0.39}$ 
        & 85.1$_{\pm0.46}$ 
        & 1152.3$_{\pm10.4}$ 
        & 55.8$_{\pm0.25}$ 
        & 46.1$_{\pm0.33}$ 
        & 0.63 
        & 10.71 \\
        & Parallel Adapter
        & 58.8$_{\pm0.32}$ 
        & 30.5$_{\pm0.32}$ 
        & 86.4$_{\pm0.46}$ 
        & 1218.7$_{\pm10.4}$ 
        & 58.0$_{\pm0.22}$ 
        & 48.8$_{\pm0.25}$ 
        & 0.72 
        & 9.75 \\
        & P-Tuning v2
        & 54.6$_{\pm0.29}$ 
        & 27.8$_{\pm0.32}$ 
        & 82.5$_{\pm0.43}$ 
        & 1114.7$_{\pm10.9}$ 
        & 53.4$_{\pm0.20}$ 
        & 45.9$_{\pm0.25}$ 
        & 0.82 
        & 8.50 \\
        & \textbf{S-MFT Attn (Ours)} 
        & 60.5$_{\pm0.24}$ 
        & 31.2$_{\pm0.30}$ 
        & \textcolor[RGB]{75,115,210}{\textbf{87.4}}$_{\pm0.45}$ 
        & 1348.8$_{\pm8.6}$ 
        & \textcolor[RGB]{235,85,75}{\textbf{63.5}}$_{\pm0.24}$ 
        & 50.3$_{\pm0.21}$ 
        & \textcolor[RGB]{235,85,75}{\textbf{0.53}} 
        & \textcolor[RGB]{235,85,75}{\textbf{7.21}} \\
        & \textbf{S-MFT MLP (Ours)} 
        & \textcolor[RGB]{75,115,210}{\textbf{61.4}}$_{\pm0.26}$ 
        & \textcolor[RGB]{235,85,75}{\textbf{31.9}}$_{\pm0.31}$ 
        & \textcolor[RGB]{75,115,210}{\textbf{87.4}}$_{\pm0.47}$ 
        & \textcolor[RGB]{235,85,75}{\textbf{1402.6}}$_{\pm9.8}$ 
        & \textcolor[RGB]{75,115,210}{\textbf{62.4}}$_{\pm0.22}$ 
        & \textcolor[RGB]{75,115,210}{\textbf{51.1}}$_{\pm0.21}$ 
        & \textcolor[RGB]{75,115,210}{\textbf{0.57}} 
        & \textcolor[RGB]{75,115,210}{\textbf{8.06}} \\
        & \textbf{S-MFT Both (Ours)} 
        & \textcolor[RGB]{235,85,75}{\textbf{61.6}}$_{\pm0.25}$ 
        & \textcolor[RGB]{235,85,75}{\textbf{31.9}}$_{\pm0.28}$ 
        & \textcolor[RGB]{235,85,75}{\textbf{87.6}}$_{\pm0.49}$ 
        & \textcolor[RGB]{75,115,210}{\textbf{1361.8}}$_{\pm9.1}$ 
        & 61.2$_{\pm0.20}$ 
        & \textcolor[RGB]{235,85,75}{\textbf{52.3}}$_{\pm0.22}$ 
        & 0.58 
        & 8.16 \\
        \midrule
        \multirow{10}{*}{\rotatebox[origin=c]{90}{\textbf{Phi-2-2.7B}}}
        & Zero-Shot
        & 2.3 
        & 25.7 
        & 50.1 
        & \multicolumn{1}{c}{87.4} 
        & 9.8 
        & 1.5 
        & - 
        & - \\
        & FFT
        & 60.5$_{\pm0.37}$ 
        & 38.2$_{\pm0.32}$ 
        & 86.4$_{\pm0.45}$ 
        & 1413.9$_{\pm9.6}$ 
        & \textcolor[RGB]{75,115,210}{\textbf{71.9}}$_{\pm0.29}$ 
        & 56.5$_{\pm0.22}$ 
        & 0.86 
        & 13.53 \\
        & LoRA
        & 59.8$_{\pm0.34}$ 
        & 37.1$_{\pm0.22}$ 
        & 88.1$_{\pm0.49}$ 
        & 1382.3$_{\pm9.7}$ 
        & 71.1$_{\pm0.20}$ 
        & 53.7$_{\pm0.25}$ 
        & 0.67 
        & 8.15 \\
        & DoRA
        & 61.0$_{\pm0.31}$ 
        & 37.9$_{\pm0.26}$ 
        & 88.2$_{\pm0.41}$ 
        & 1395.2$_{\pm8.8}$ 
        & 71.5$_{\pm0.23}$ 
        & 55.2$_{\pm0.27}$ 
        & 0.65 
        & 8.65 \\
        & QLoRA
        & 58.3$_{\pm0.39}$ 
        & 33.9$_{\pm0.28}$ 
        & 85.7$_{\pm0.47}$ 
        & 1354.6$_{\pm10.3}$ 
        & 69.4$_{\pm0.28}$ 
        & 56.9$_{\pm0.37}$ 
        & 0.72 
        & 11.18 \\
        & Parallel Adapter
        & 60.3$_{\pm0.34}$ 
        & 36.8$_{\pm0.20}$ 
        & 86.8$_{\pm0.45}$ 
        & 1388.7$_{\pm9.7}$ 
        & 70.8$_{\pm0.27}$ 
        & 54.4$_{\pm0.24}$ 
        & 0.69 
        & 8.42 \\
        & P-Tuning v2
        & 57.0$_{\pm0.33}$ 
        & 31.5$_{\pm0.39}$ 
        & 84.5$_{\pm0.47}$ 
        & 1280.6$_{\pm9.6}$ 
        & 66.3$_{\pm0.28}$ 
        & 51.0$_{\pm0.31}$ 
        & 0.81 
        & 7.84 \\
        & \textbf{S-MFT Attn(Ours)} 
        & 60.3$_{\pm0.24}$ 
        & 37.8$_{\pm0.27}$ 
        & \textcolor[RGB]{75,115,210}{\textbf{88.3}}$_{\pm0.45}$ 
        & 1403.9$_{\pm9.3}$ 
        & \textcolor[RGB]{235,85,75}{\textbf{72.4}}$_{\pm0.20}$ 
        & 56.1$_{\pm0.25}$ 
        & \textcolor[RGB]{235,85,75}{\textbf{0.46}} 
        & \textcolor[RGB]{235,85,75}{\textbf{5.21}} \\
        & \textbf{S-MFT MLP (Ours)} 
        & \textcolor[RGB]{75,115,210}{\textbf{61.4}}$_{\pm0.25}$ 
        & \textcolor[RGB]{75,115,210}{\textbf{38.6}}$_{\pm0.25}$ 
        & 88.0$_{\pm0.47}$ 
        & \textcolor[RGB]{235,85,75}{\textbf{1468.3}}$_{\pm9.1}$ 
        & 71.7$_{\pm0.23}$ 
        & \textcolor[RGB]{235,85,75}{\textbf{58.8}}$_{\pm0.31}$ 
        & \textcolor[RGB]{75,115,210}{\textbf{0.54}} 
        & \textcolor[RGB]{75,115,210}{\textbf{7.33}} \\
        & \textbf{S-MFT Both(Ours)} 
        & \textcolor[RGB]{235,85,75}{\textbf{61.9}}$_{\pm0.26}$ 
        & \textcolor[RGB]{235,85,75}{\textbf{39.1}}$_{\pm0.28}$ 
        & \textcolor[RGB]{235,85,75}{\textbf{88.8}}$_{\pm0.45}$ 
        & \textcolor[RGB]{75,115,210}{\textbf{1444.5}}$_{\pm9.0}$ 
        & \textcolor[RGB]{235,85,75}{\textbf{72.4}}$_{\pm0.21}$ 
        & \textcolor[RGB]{75,115,210}{\textbf{57.2}}$_{\pm0.28}$ 
        & 0.64 
        & 8.12 \\
        \bottomrule
    \end{tabular}
    }
    \label{tab:exps_main}
    \vspace{-6mm}
\end{table*}

\subsection{Training Objective}
MFT uses the same token-level negative log-likelihood as FFT, with the effective weights replaced by $W\odot M$:
\begin{equation}\label{eq:mft-revised}
\scalebox{0.9}{$
\mathcal{L}(U_m;M)
=-\frac{1}{L}\sum_{i=1}^{L}
\log P\!\left(u_m^i\mid u_m^{<i};W\odot M\right)
$}.
\end{equation}
where $U_m=\{u_m^1,\ldots,u_m^L\}$ is the target token sequence. The pre-trained weights $W$ are fixed, and we optimize the mask scores.

\subsection{Theoretical Grounding}
\label{sec:theory}
Condition on pre-trained weights $W_0$ fixed before an i.i.d.\ downstream sample $\mathcal S$ of size $n$, and let $R$ and $\widehat R_{\mathcal S}$ denote population and empirical risk for a bounded loss $\ell\in[0,1]$. On the $d$ mask-eligible weights, compare MFT and FFT at the same effective update radius $r$, and write $\mathbb{B}_2(r)=\{\Delta W:\|\Delta W\|_2\le r\}$:
\begin{equation}
\scalebox{0.9}{$
\begin{aligned}
\mathcal U_{\mathrm{MFT}}(r)
&=\{W_0\odot(M-\mathbf 1):M\in[0,1]^d\}
\cap\mathbb{B}_2(r)\\
&\subseteq\mathcal U_{\mathrm{FFT}}(r)=\mathbb{B}_2(r),\\
\mathcal H_{\mathrm{MFT}}(r)
&\subseteq\mathcal H_{\mathrm{FFT}}(r)
\end{aligned}
$}.
\label{eq:class-inclusion}
\end{equation}
where
$\mathcal H_A(r)=
\{x\mapsto f(x;W_0+\Delta W):
\Delta W\in\mathcal U_A(r)\}$.
The inclusion follows because every MFT update obeys
$\Delta W_iW_{0,i}\le0$ and
$|\Delta W_i|\le|W_{0,i}|$.
Monotonicity of empirical Rademacher complexity then gives
\begin{equation}
\widehat{\mathfrak R}_n
\!\left(\ell\circ\mathcal H_{\mathrm{MFT}}(r)\right)
\le
\widehat{\mathfrak R}_n
\!\left(\ell\circ\mathcal H_{\mathrm{FFT}}(r)\right).
\label{eq:rad-inclusion}
\end{equation}
Hence, the standard Rademacher generalization bound~\citep{bartlett2002rademacher} gives MFT a no-larger complexity term at a matched effective-weight budget. This establishes MFT's capacity-control advantage: its structure controls statistical complexity while the experiments show that it preserves or improves downstream task performance. The compression analysis below strengthens this ordering with an explicit sufficient condition for a strictly tighter MFT bound. The full risk statement and proof are in Appendix~\ref{sec:appendix_theory}.

\begin{table}[t]
    \caption{
    Comparison of MFT and strong mask selection baselines.
    }
    \vspace{-2mm}
    \centering
    \setlength{\tabcolsep}{8pt}
    \adjustbox{width=0.85\columnwidth}{
    \begin{tabular}{llcccccc}
        \toprule
        \multicolumn{2}{c}{\multirow{2}{*}{\textbf{Method}}} 
        & \multicolumn{6}{c}{\textbf{Evaluation Tasks}} \\
        \cmidrule(lr){3-8}
        \multicolumn{2}{l}{} 
        & \multicolumn{1}{c}{\textbf{GQA} {↑}} 
        & \multicolumn{1}{c}{\textbf{MMMU} {↑}} 
        & \multicolumn{1}{c}{\textbf{POPE} {↑}} 
        & \multicolumn{1}{c}{\textbf{MME} {↑}} 
        & \multicolumn{1}{c}{\textbf{SQA-Image} {↑}} 
        & \multicolumn{1}{c}{\textbf{TextVQA} {↑}} \\
        \midrule

        \multirow{8}{*}{\rotatebox[origin=c]{90}{\textbf{Qwen2.5-0.5B}}}
        & Zero-Shot
        & 5.7 
        & 22.6 
        & 49.9
        & 93.7
        & 3.2 
        & 9.0 \\
        & FFT
        & 57.8 
        & 32.8 
        & 86.4
        & 1278.5
        & 58.2 
        & 47.6 \\
        \cmidrule(lr){2-8}
        & Wanda 0.05\%
        & 58.2 
        & 32.4
        & 86.1
        & 1290.5
        & 58.8 
        & 47.5 \\
        & SparseGPT 0.05\%
        & 58.0 
        & 32.2
        & 85.9
        & 1286.2
        & 58.6 
        & 47.4 \\
        & H-MFT 0.05\% 
        & \textbf{59.1}
        & \textbf{33.8}
        & \textbf{86.6}
        & \textbf{1345.7}
        & \textbf{59.4} 
        & \textbf{48.0} \\
        \cmidrule(lr){2-8}
        & H-MFT 0.01\% 
        & 56.1
        & 31.7
        & 85.9
        & 1302.1
        & 57.9 
        & 43.9 \\
        & H-MFT 0.1\% 
        & 53.0
        & 31.4
        & 85.3
        & 1212.4
        & 57.9 
        & 44.8 \\
        & S-MFT 
        & 59.0 
        & 34.6
        & 87.2
        & 1342.1
        & 59.5 
        & 48.8 \\

        \midrule
        \multirow{8}{*}{\rotatebox[origin=c]{90}{\textbf{TinyLLaMA-1.1B}}}
        & Zero-Shot
        & 1.0 
        & 26.1
        & 49.8
        & 211.8
        & 4.0 
        & 6.8 \\
        & FFT
        & 56.9 
        & 29.8
        & 86.2
        & 1165.3
        & 59.1 
        & 47.2 \\
        \cmidrule(lr){2-8}
        & Wanda 1.0\% 
        & 56.5 
        & 30.5
        & 86.4
        & 1175.2
        & 58.8 
        & 47.6 \\
        & SparseGPT 1.0\% 
        & 56.3 
        & 30.3
        & 86.2
        & 1171.5
        & 58.6 
        & 47.4 \\
        & H-MFT 1.0\% 
        & \textbf{57.5} 
        & \textbf{31.6}
        & \textbf{86.9}
        & \textbf{1191.9}
        & \textbf{59.5} 
        & \textbf{48.2} \\
        \cmidrule(lr){2-8}
        & H-MFT 0.5\% 
        & 56.2 
        & 28.1
        & 86.7
        & 1174.7
        & 54.9 
        & 43.1 \\
        & H-MFT 1.5\% 
        & 51.2 
        & 28.2
        & 84.1
        & 1112.4
        & 57.1 
        & 46.2 \\
        & S-MFT 
        & 57.8 
        & 32.3
        & 86.8
        & 1196.1
        & 60.7 
        & 48.4 \\

        \midrule
        \multirow{8}{*}{\rotatebox[origin=c]{90}{\textbf{Gemma2-2B}}}
        & Zero-Shot
        & 2.6 
        & 26.4
        & 56.6
        & 586.2
        & 2.8 
        & 5.6 \\
        & FFT
        & 60.2 
        & 30.8
        & 86.8
        & 1318.4
        & 59.7 
        & 50.2 \\
        \cmidrule(lr){2-8}
        & Wanda 3\% 
        & 60.6 
        & 30.4
        & 86.4
        & 1325.6
        & 59.8 
        & 50.6 \\
        & SparseGPT 3\% 
        & 60.4 
        & 30.2
        & 86.3
        & 1322.1
        & 59.5 
        & 50.4 \\
        & H-MFT 3\% 
        & \textbf{61.5} 
        & \textbf{31.8}
        & \textbf{86.9}
        & \textbf{1351.5}
        & \textbf{60.5} 
        & \textbf{51.4} \\
        \cmidrule(lr){2-8}
        & H-MFT 1\% 
        & 55.1
        & 31.4
        & 85.9
        & 1272.3
        & 59.0 
        & 47.2 \\
        & H-MFT 5\% 
        & 55.4 
        & 30.7
        & 85.5
        & 1242.1
        & 60.0 
        & 48.9 \\
        & S-MFT 
        & 61.4 
        & 32.1
        & 87.1
        & 1362.7
        & 60.4 
        & 52.0 \\

        \midrule
        \multirow{8}{*}{\rotatebox[origin=c]{90}{\textbf{Phi-2-2.7B}}}
        & Zero-Shot
        & 2.3 
        & 25.7
        & 50.1
        & 87.4
        & 9.8 
        & 1.5 \\
        & FFT
        & 61.2 
        & 37.5
        & 87.2
        & 1398.6
        & 71.4 
        & 56.1 \\
        \cmidrule(lr){2-8}
        & Wanda 0.1\% 
        & 61.6 
        & 37.3
        & 87.1
        & 1408.2
        & 71.6 
        & 56.0 \\
        & SparseGPT 0.1\% 
        & 61.4 
        & 37.1
        & 87.0
        & 1404.5
        & 71.4 
        & 55.8 \\
        & H-MFT 0.1\% 
        & \textbf{62.1} 
        & \textbf{38.4}
        & \textbf{87.7}
        & \textbf{1427.0}
        & \textbf{72.1} 
        & \textbf{56.6} \\
        \cmidrule(lr){2-8}
        & H-MFT 0.05\% 
        & 61.1 
        & 37.1
        & 87.3
        & 1393.3
        & 71.0 
        & 56.7 \\
        & H-MFT 0.3\% 
        & 60.7 
        & 35.9
        & 86.8
        & 1345.6
        & 69.8 
        & 55.4 \\
        & S-MFT 
        & 62.1 
        & 38.8
        & 88.0
        & 1423.5
        & 72.6 
        & 57.0 \\

        \bottomrule
    \end{tabular}
    }
    \vspace{-6mm}
    \label{tab:hard-mask}
\end{table}

Eq.~\ref{eq:rad-inclusion} establishes that MFT's complexity is upper-bounded by FFT's. We next derive an explicit sufficient condition for a strict separation when a learned soft mask $M$ admits a short code. Let $\widetilde M$ leave all but $s$ mask coordinates equal to one and encode each selected coordinate with a fixed, data-independent $b$-bit alphabet. Define its worst-case loss distortion as
\begin{equation}
\scalebox{0.75}{$
\alpha(M,\widetilde M)
=\sup_z
\left|
\ell(f(\cdot;W_0\odot M),z)
-\ell(f(\cdot;W_0\odot\widetilde M),z)
\right|
$}.
\end{equation}
Since there are at most $\binom{d}{s}2^{bs}$ such codes, a finite-class bound gives, with probability at least $1-\delta$,
\begin{equation}
\scalebox{0.9}{$
\begin{aligned}
R(h_M)
\le{}&\widehat R_{\mathcal S}(h_M)
+2\alpha(M,\widetilde M)\\
&+\sqrt{
\frac{
\ln \binom{d}{s}
+bs\ln2
+\ln(1/\delta)}
{2n}}
\end{aligned}
$}.
\label{eq:soft-compression-bound}
\end{equation}
The analogous numerator for a dense $b_F$-bit FFT code is $b_Fd\ln2$. Thus, MFT has a strictly smaller complexity contribution whenever
$\ln \binom{d}{s}+bs\ln2<b_Fd\ln2$;
combining this coding advantage with preserved empirical fit and small mask distortion yields a strictly tighter complete risk bound for MFT. The result provides an explicit, empirically checkable mechanism through which structural selection improves generalization over dense FFT updates.

For the score-to-weight map $\Phi(S)=W_0\odot\sigma_T(S)$,
\begin{equation}
\|\Phi(S_1)-\Phi(S_2)\|_2
\le
\frac{\|W_0\|_\infty}{4T}
\|S_1-S_2\|_2.
\label{eq:revised-lipschitz}
\end{equation}
Moreover, the chain rule gives
\begin{equation}
\scalebox{0.9}{$
\begin{aligned}
&\nabla_S\mathcal L
=
\left(\nabla_{W'}\mathcal L\odot W_0\right)
\odot\sigma_T'(S),\\
&\|\nabla_S\mathcal L\|_2
\le
\frac{\|W_0\|_\infty}{4T}
\|\nabla_{W'}\mathcal L\|_2
\end{aligned}
$}.
\label{eq:gradient-sensitivity}
\end{equation}
These inequalities explicitly bound score-to-weight sensitivity. The sigmoid derivative selectively damps score updates on saturated coordinates, providing a stable mechanism for consolidating learned structural selections. The complete assumptions, post-selection corrections, H-MFT and FFT special cases, and all proofs are in Appendix~\ref{sec:appendix_theory}.

\section{Experiments}
\label{sec:experiments}
We apply our Mask Fine-tuning (MFT) method on Vision-Language Models (VLMs) across different vision and language backbones and evaluate on diverse vision-language tasks. We also provide ablation studies and analyses from various perspectives to gain deeper insights into our method.

\subsection{Experimental Setup}
\textbf{Vision Encoders.} We follow the mainstream VLM training paradigm, in which the vision encoder is frozen, and the projector and language model are trained~\citep {liu2023llava,zhou2024tinyllava}. Accordingly, we adopt CLIP~\citep{radford2021learningtransferablevisualmodels} and SigLIP~\citep{zhai2023sigmoidlosslanguageimage} in our experiments.

\begin{figure*}[t]
    \centering
    \includegraphics[width=1\linewidth]{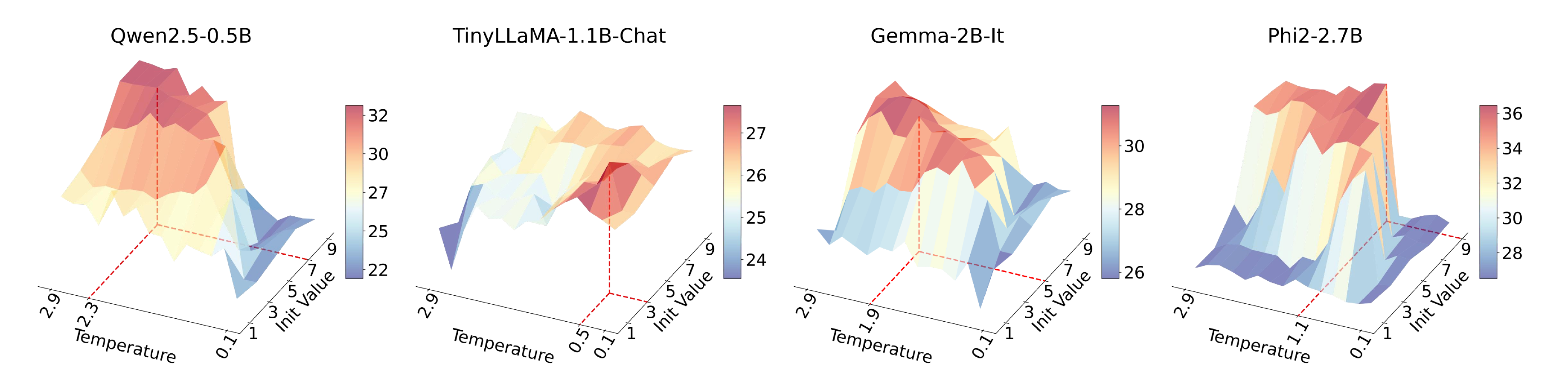}
    \vspace{-8mm}
    \caption{
    Hyperparameter sensitivity of S-MFT with respect to the initial score value ($S_{init}$) and temperature ($T$) across four language backbones. Each surface plot shows the MMMU performance under different hyperparameter combinations. The \textcolor[RGB]{235,85,75}{red dashed line} indicates the configuration selected for full training.
    }
    \label{fig:init-temp-ablation}
\end{figure*}

\begin{table*}[t]
    \centering
    \renewcommand{\arraystretch}{0.9}
    \setlength{\tabcolsep}{3pt}
    
    \begin{minipage}[t]{0.48\textwidth}
        \centering
        \captionof{table}{Ablation of different vision towers.}
        \vspace{-2mm}
        \label{tab:vision-tower}
        \adjustbox{width=\textwidth}{
        \begin{tabular}{lcccccc}
            \toprule
            \textbf{Method} 
            & \multicolumn{1}{c}{GQA {↑}} 
            & \multicolumn{1}{c}{MMMU {↑}} 
            & \multicolumn{1}{c}{POPE {↑}} 
            & \multicolumn{1}{c}{MME {↑}} 
            & \multicolumn{1}{c}{SQA-Image {↑}} 
            & \multicolumn{1}{c}{TextVQA {↑}} \\
            \midrule
            \multicolumn{7}{c}{\textbf{Qwen2.5-0.5B + CLIP-ViT-Large}} \\
            \midrule
            FFT 
            & 55.7 
            & 34.2 
            & 85.3 
            & 1197.9 
            & 57.6 
            & 44.2 \\
            LoRA 
            & 55.1 
            & 32.7 
            & 85.2 
            & 1196.7
            & 57.5 
            & 43.7 \\
            S-MFT 
            & \textbf{56.8}
            & \textbf{35.2}
            & \textbf{86.3}
            & \textbf{1238.4}
            & \textbf{58.9}
            & \textbf{45.2} \\
            \midrule
            \multicolumn{7}{c}{\textbf{TinyLlama-1.1B + CLIP-ViT-Large}} \\
            \midrule
            FFT 
            & 58.0 
            & 27.9 
            & 85.5 
            & 1284.6 
            & 59.9 
            & 46.3 \\
            LoRA 
            & 57.8
            & 27.5
            & 85.4
            & 1278.4
            & 59.3
            & 45.9 \\
            S-MFT 
            & \textbf{59.2} 
            & \textbf{29.0} 
            & \textbf{86.5} 
            & \textbf{1319.0} 
            & \textbf{61.0} 
            & \textbf{47.2} \\
            \midrule
            \multicolumn{7}{c}{\textbf{Gemma-2B + CLIP-ViT-Large}} \\
            \midrule
            FFT 
            & 56.4 
            & 32.3
            & 84.4
            & 1194.1
            & 60.0
            & 45.8 \\
            LoRA 
            & 57.2 
            & 28.4 
            & 83.9 
            & 1249.7 
            & 57.6 
            & 45.9 \\
            S-MFT 
            & \textbf{58.4} 
            & \textbf{33.4} 
            & \textbf{85.8} 
            & \textbf{1254.4} 
            & \textbf{60.9} 
            & \textbf{47.0} \\
            \midrule
            \multicolumn{7}{c}{\textbf{Phi-2-2.7B + CLIP-ViT-Large}} \\
            \midrule
            FFT 
            & 59.4
            & 36.3
            & 86.8
            & 1448.6
            & 71.2
            & 53.4 \\
            LoRA 
            & 58.7
            & 35.8
            & 86.6
            & 1437.1
            & 70.9
            & 53.3 \\
            S-MFT 
            & \textbf{60.3}
            & \textbf{37.3}
            & \textbf{87.9}
            & \textbf{1477.9}
            & \textbf{72.1}
            & \textbf{55.0} \\
            \bottomrule
        \end{tabular}
        }
    \end{minipage}
    \hfill
    \begin{minipage}[t]{0.48\textwidth}
            \centering
            \captionof{table}{Scaling study of S-MFT on VLMs.}
            \vspace{-2mm}
            \label{tab:scale-up}
            \adjustbox{width=\textwidth}{
            \begin{tabular}{lcccccc}
                \toprule
                \textbf{Method} 
                & \multicolumn{1}{c}{GQA {↑}} 
                & \multicolumn{1}{c}{MMMU {↑}} 
                & \multicolumn{1}{c}{POPE {↑}} 
                & \multicolumn{1}{c}{MME {↑}} 
                & \multicolumn{1}{c}{SQA-Image {↑}} 
                & \multicolumn{1}{c}{TextVQA {↑}} \\
                \midrule
                \multicolumn{7}{c}{\textbf{Qwen2.5-7B + SigLIP-ViT-Large}} \\
                \midrule
                FFT 
                & 64.7
                & 43.8
                & 87.7
                & 1558.3
                & 76.4
                & 64.0 \\
                LoRA 
                & 63.9
                & 42.6
                & 86.8
                & 1512.4
                & 74.5
                & 63.7 \\
                S-MFT 
                & \textbf{65.6}
                & \textbf{44.7}
                & \textbf{88.6}
                & \textbf{1588.1}
                & \textbf{77.7}
                & \textbf{65.3} \\
                \midrule
                \multicolumn{7}{c}{\textbf{Qwen2.5-14B + SigLIP-ViT-Large}} \\
                \midrule
                FFT 
                & 65.6
                & 47.8
                & 88.5
                & 1597.2
                & 80.1
                & 66.1 \\
                LoRA 
                & 65.3
                & 46.5
                & 87.8
                & 1568.9
                & 79.1
                & 66.0 \\
                S-MFT 
                & \textbf{66.5}
                & \textbf{48.9}
                & \textbf{89.6}
                & \textbf{1627.5}
                & \textbf{81.2}
                & \textbf{67.2} \\
                \midrule
                \multicolumn{7}{c}{\textbf{LLaMA2-7B + CLIP-ViT-Large}} \\
                \midrule
                FFT 
                & 61.7
                & 40.2
                & 87.2
                & 1506.6
                & 71.8
                & 57.7 \\
                LoRA 
                & 61.9
                & 38.8
                & 87.6
                & 1490.5
                & 71.3
                & 56.5 \\
                S-MFT 
                & \textbf{62.7}
                & \textbf{41.3}
                & \textbf{88.3}
                & \textbf{1535.9}
                & \textbf{72.9}
                & \textbf{58.7} \\
                \midrule
                \multicolumn{7}{c}{\textbf{LLaMA2-13B + CLIP-ViT-Large}} \\
                \midrule
                FFT 
                & 63.5
                & 43.6
                & 88.0
                & 1545.2
                & 74.4
                & 59.8 \\
                LoRA 
                & 63.1
                & 42.4
                & 87.9
                & 1528.8
                & 73.8
                & 59.0 \\
                S-MFT 
                & \textbf{64.4}
                & \textbf{44.7}
                & \textbf{88.9}
                & \textbf{1578.4}
                & \textbf{75.5}
                & \textbf{61.0} \\
                \bottomrule
            \end{tabular}
            }
    \end{minipage}
    \vspace{-4mm}
\end{table*}

\noindent\textbf{Language Models.} 
We primarily explore and validate MFT on four capable but compact language backbones: Qwen2.5-0.5B~\citep{qwen2025qwen25technicalreport}, TinyLLaMA-1.1B~\citep{zhang2024tinyllama}, Gemma-2B~\citep{team2024gemma}, and Phi-2-2.7B~\citep{gunasekar2023textbooks}.
We further validate the transferability of the discovered structural selection principles through targeted scale-up experiments on larger mainstream language backbones.

\noindent\textbf{Baselines \& Datasets.} We follow the settings of LLaVA~\citep{liu2023improvedllava} and TinyLLaVA~\citep{zhou2024tinyllava}, which serve as our strong fine-tuning baselines. Specifically, we use exactly the same training data as in their setups and evaluate our models across multiple downstream tasks, including GQA~\citep{ainslie2023gqatraininggeneralizedmultiquery}, MMMU~\citep{yue2023mmmu}, POPE~\citep{Li-hallucination-2023}, MME~\citep{fu2025mmecomprehensiveevaluationbenchmark}, SQA~\citep{lu2022learn}, and TextVQA~\citep{singh2019towards}. 
We also compare MFT against several representative PEFT approaches, including LoRA~\citep{hu2021loralowrankadaptationlarge}, DoRA~\citep{liu2024doraweightdecomposedlowrankadaptation}, QLoRA~\citep{dettmers2023qloraefficientfinetuningquantized}, Parallel Adapter~\citep{he2022towards}, and P-Tuning v2~\citep{DBLP:journals/corr/abs-2110-07602}, to demonstrate its effectiveness. Additional comparisons are presented in the Appendix.

\subsection{Main Results}
We use Soft Mask Fine-Tuning (S-MFT) as the primary experimental setting and conduct extensive ablation studies to demonstrate its effectiveness, efficiency, and guidance to the binary variant.

\noindent\textbf{Comparison.} Tab.~\ref{tab:exps_main} contains our main experiment results of S-MFT, which achieves strong and consistent performance across all four backbones and six evaluation tasks. All experiments follow the TinyLLaVA~\citep{zhou2024tinyllava} setting and train for one epoch with SigLIP~\citep{zhai2023sigmoidlosslanguageimage} as the vision tower. We implement three S-MFT variants: 1) S-MFT Attn, 2) S-MFT MLP, and 3) S-MFT Both, which apply learnable masks to the attention weights, MLP weights, or both within Transformer~\citep{wolf-etal-2020-transformers} blocks.
The ``Epochs'' and ``Time'' columns report the training duration required for convergence. S-MFT consistently achieves strictly lower training costs than FFT. Within this reduced time budget, the strongest S-MFT configuration consistently surpasses FFT and PEFT methods across diverse visual-linguistic benchmarks.
Among S-MFT variants, S-MFT Both consistently yields the highest overall accuracy, while module-specific masking provides efficient operating points with further computational savings. These results establish S-MFT as a robust and efficient fine-tuning method that surpasses full tuning across diverse VLM architectures.

\begin{figure*}[t]
    \centering
    
    \begin{minipage}[t]{0.48\linewidth}
        \centering
        \includegraphics[width=\linewidth]{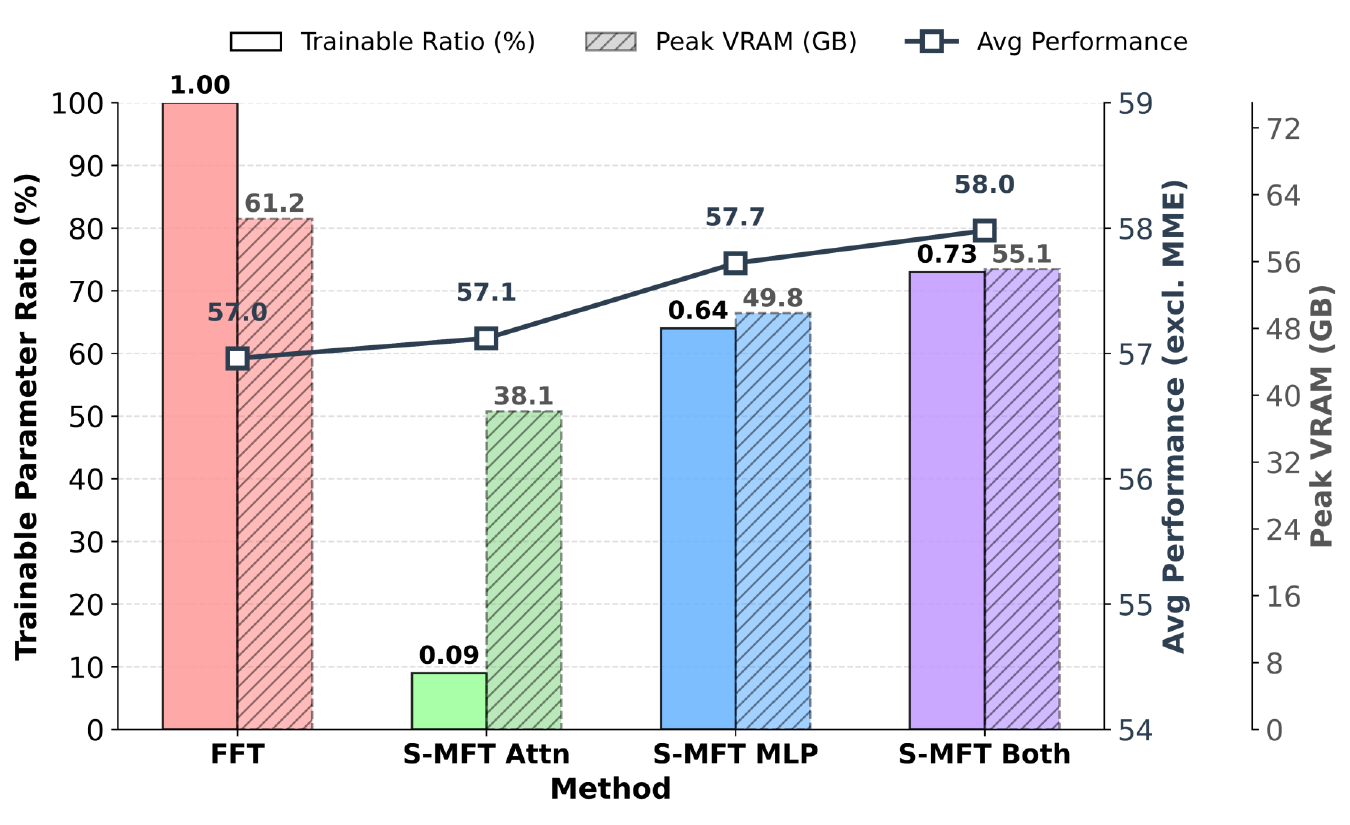}
        \vspace{-8mm}
        \caption{
        Comparison of training cost and average performance between FFT and MFT.
        }
        \label{fig:trainable_params}
    \end{minipage}
    \hfill
    \begin{minipage}[t]{0.48\linewidth}
        \centering
        \raisebox{0.5mm}{\includegraphics[width=\linewidth]{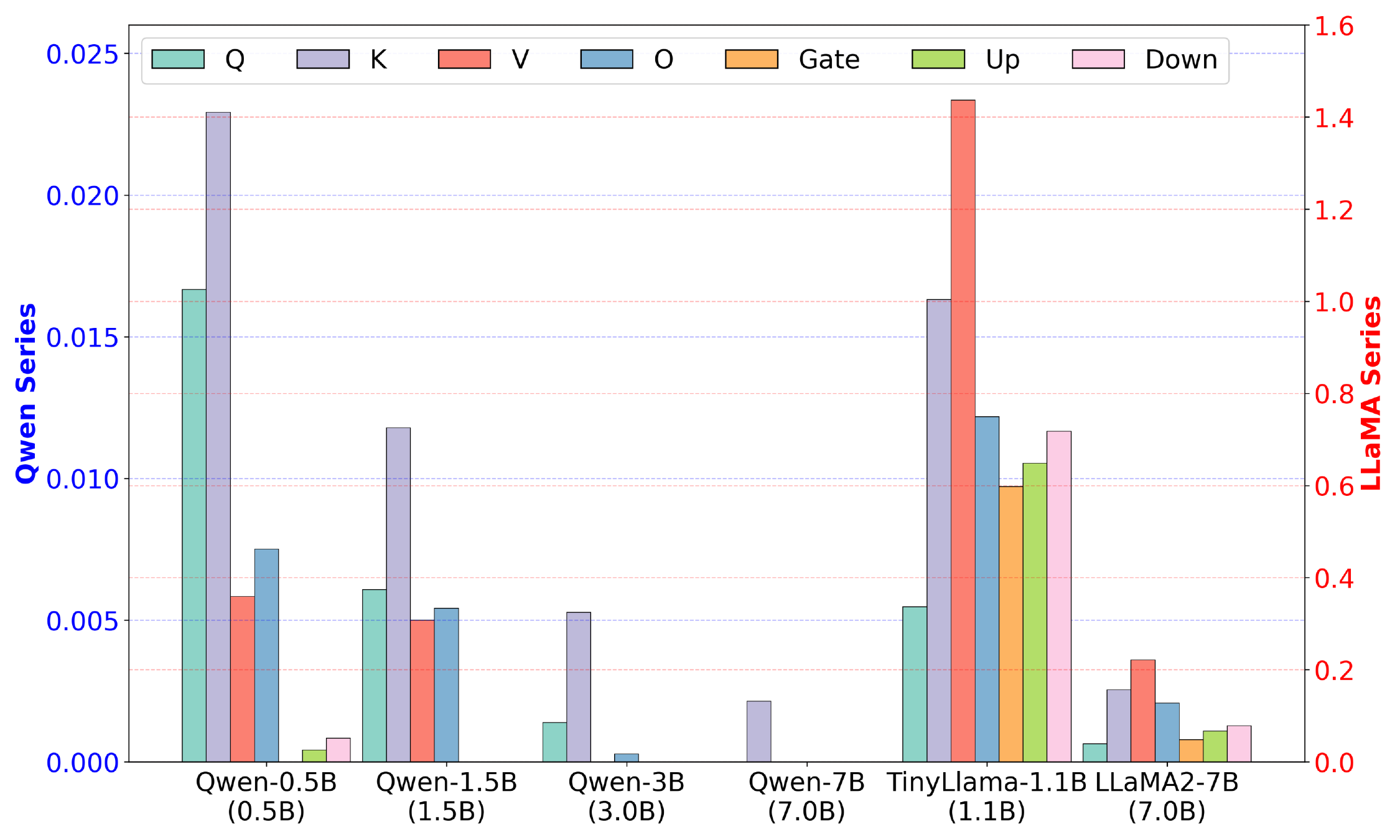}}
        \vspace{-8mm}
        \caption{
        The proportion of near-zero mask values learned by S-MFT.
        }
        \label{fig:mask-ratio}
    \end{minipage}
    \vspace{-6mm}
\end{figure*}

\noindent\textbf{H-MFT vs. Pruning.} 
In Tab.~\ref{tab:hard-mask}, H-MFT achieves the best performance when using the approximated sparsity level adaptively learned by S-MFT (Fig.~\ref{fig:mask-ratio}). A key insight of MFT is that adapting multimodal models does not require additional information. Instead, effective adaptation can be achieved by modifying only a tiny fraction of connections via mask selection. To verify that the performance gains arise from task-aware structural selection rather than generic sparsity effects, we compare H-MFT with Wanda~\citep{sun2023wanda} and SparseGPT~\citep{frantar-sparsegpt} at identical sparsities. To ensure strict optimization parity, we first derive static pruning masks for the base model and then perform sparse fine-tuning under the same multimodal training configuration as H-MFT. These task-agnostic heuristics select weights using magnitude or general activation statistics, providing controls that isolate the contribution of multimodal task supervision. H-MFT's task-aware masks consistently surpass FFT and both heuristic control groups. These results show that MFT's gains arise from learned structural selection rather than generic sparsity or regularization.

\begin{figure*}[t]
    \centering
    \includegraphics[width=1.0\linewidth]{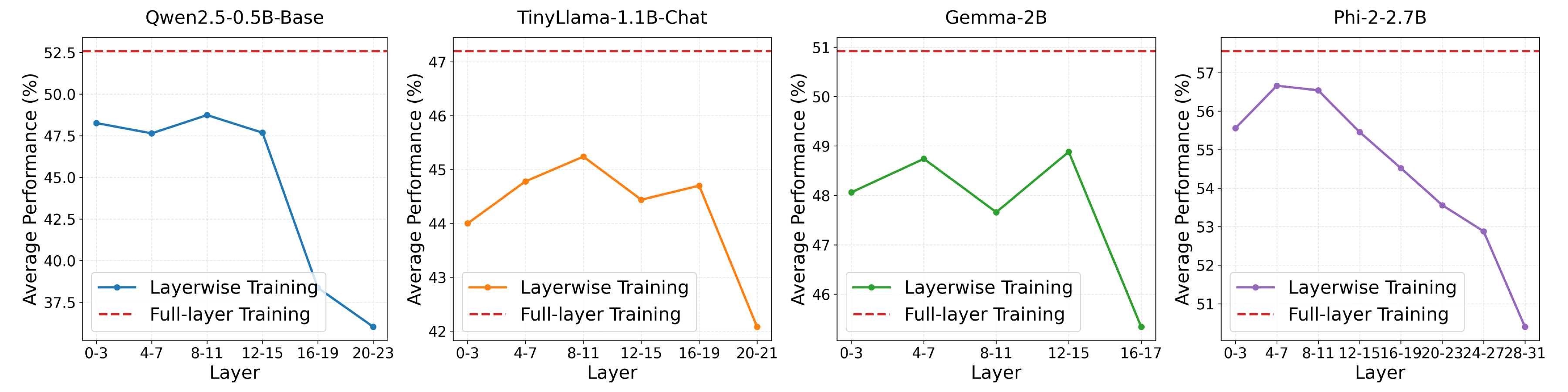}
    \vspace{-8mm}
    \caption{
    Layer-wise analysis of S-MFT across four language backbones. Each curve shows the average performance when masks are applied only to a subset of layers, compared with full-layer training (\textcolor[RGB]{235,85,75}{red dashed line}).
    }
    \label{fig:layerwise}
\end{figure*}

\subsection{Ablation Studies}
\noindent\textbf{Hyperparameters.} 
S-MFT mainly depends on two hyperparameters: the initial score $S_{init}$ and the temperature $T$. 
Before full-scale training, we perform a lightweight grid search on 10\% of the training data to evaluate each model's sensitivity to these two hyperparameters (Fig.~\ref{fig:init-temp-ablation}). 
As shown in Fig.~\ref{fig:init-temp-ablation}, the performance surfaces show broad, continuous high-performance regions rather than sharp peaks, indicating that S-MFT is robust to variations in both $S_{init}$ and $T$. 
We therefore select the optimal configuration as the default setting for full-dataset training. 
Concretely, we use ($S_{init}$-$T$) values of 7.0-2.3, 3.0-0.5, 5.0-1.9, and 9.0-1.1 for Qwen2.5-0.5B, TinyLLaMA-1.1B, Gemma-2B, and Phi-2-2.7B, respectively.

\noindent\textbf{Vision Towers.}
We evaluate S-MFT with CLIP~\citep{radford2021learningtransferablevisualmodels} in addition to the default SigLIP~\citep{zhai2023sigmoidlosslanguageimage} tower to validate cross-encoder generality in Tab.~\ref{tab:vision-tower}. S-MFT consistently outperforms FFT and LoRA across these vision-language combinations, demonstrating robust generalization across vision towers.

\noindent\textbf{Scale-Up.} 
To demonstrate the scalability of our method, we extend S-MFT to larger language backbones in the Qwen2.5 and LLaMA2 families. Models within each family exhibit similar mask distributions (Fig.~\ref{fig:mask-ratio}), enabling direct zero-shot transfer of $(S_{init},T)$ from the corresponding sub-billion-parameter proxy model.
Without any additional hyperparameter search on the larger models, S-MFT consistently outperforms FFT and LoRA across the reported downstream benchmarks. These results establish strong architecture-family transferability and amortize the initial profiling cost, yielding a practical ``proxy-tuning'' paradigm in which selection priors discovered on a compact model transfer directly to substantially larger models.

\noindent\textbf{Layer-wise Ablation.} We conduct a layer-wise sensitivity study with 10\% of the training data to examine where mask learning is most effective (Fig.~\ref{fig:layerwise}). Across all four backbones, focused masking of mid-level layers achieves the highest average performance and approaches full-layer training. This concentration of gains identifies mid-level reconfiguration as the most efficient adaptation region, providing a principled route to further reduce training cost through targeted layer selection.

\section{Analysis}
\label{sec:analysis}
\noindent\textbf{Training Memory.} 
Unlike traditional PEFT, designed for extreme parameter reduction at the cost of narrow informational bottlenecks, MFT is fundamentally a high-capacity, structural selection alternative to FFT. During the backward pass, MFT preserves the dense forward activations required to compute score gradients through the chain rule, retaining the model's full cross-modal representational capacity.
MFT obtains its memory advantage by freezing the pre-trained weights and eliminating their optimizer states; optimizer storage is confined to trainable scores in the selected masked modules.
Empirically, in Fig.~\ref{fig:trainable_params}, on a Qwen2.5-0.5B backbone, FFT consumes 61.2 GB of peak VRAM, while S-MFT Both uses 55.1 GB while exploring a broad 73\% trainable score space. S-MFT Attn provides an especially efficient operating point, which optimizes only 9\% of the parameter space, reducing peak VRAM to 38.1 GB, avoiding the FFT out-of-memory regime, and preserving dense-level multimodal alignment.

\noindent\textbf{Mask structure.}
Fig.~\ref{fig:mask-ratio} reports the proportion of learned mask values below $0.1$ for each projection. Together with the task-aware pruning controls in Tab.~\ref{tab:hard-mask}, the differentiated, architecture-specific attenuation patterns provide evidence that S-MFT learns structured selections rather than applying uniform shrinkage. These endpoint statistics identify candidate coordinates for the compact approximations analyzed by Eq.~\ref{eq:soft-compression-bound}, providing empirical inputs for evaluating MFT's compression-based generalization advantage.

\begin{wrapfigure}{r}{0.6\columnwidth}
    \centering
    \vspace{-10pt}
    \includegraphics[width=\linewidth]{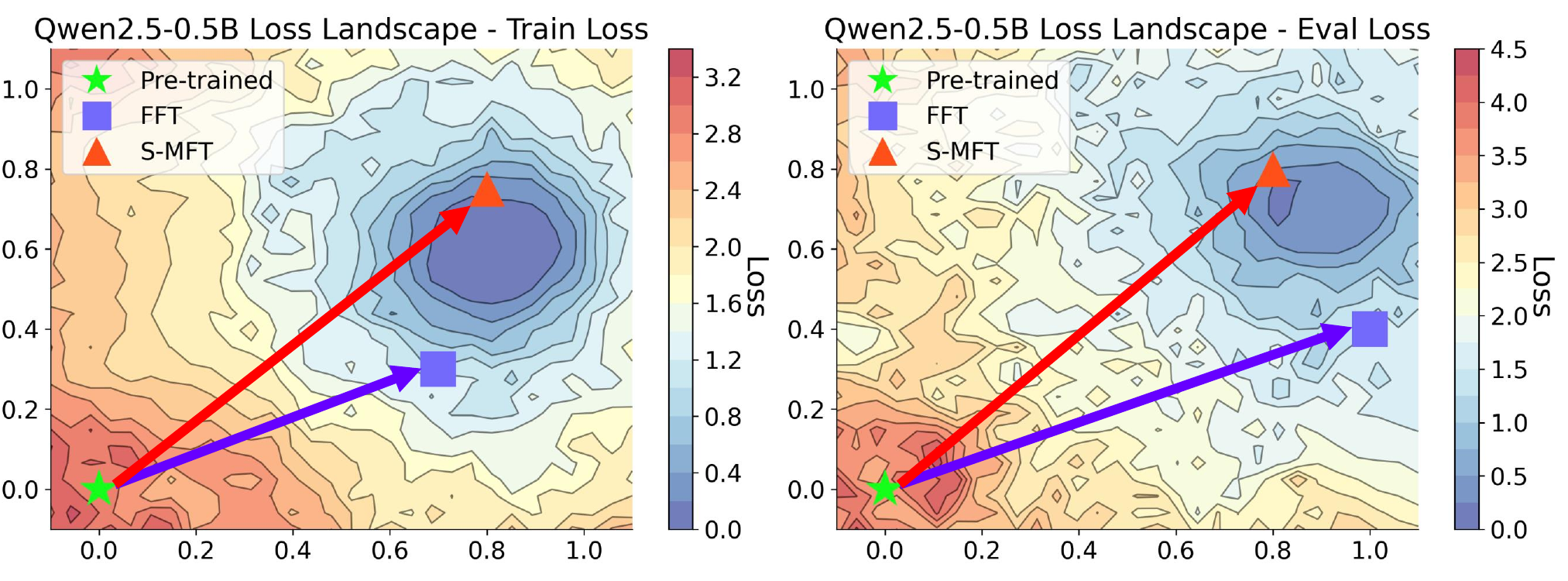}
    \vspace{-16pt}
    \caption{
    Loss landscape of the Qwen2.5-0.5B backbone.
    }
    \label{fig:landscape}
    \vspace{-8pt}
\end{wrapfigure}
\noindent\textbf{Loss landscape.} Fig.~\ref{fig:landscape} complements the formal analysis with a filter-normalized two-dimensional view of the loss near the learned solutions. The S-MFT endpoint occupies a visibly wider low-loss region than FFT, providing qualitative evidence of stronger local perturbation robustness.
Following the visualization procedure of~\citep{NEURIPS2018_a41b3bb3}, we use filter-normalized directions to control simple scale effects. The broader S-MFT basin provides complementary empirical evidence that MFT combines the controlled hypothesis complexity established in Sec.~\ref{sec:theory} with stronger local robustness along the displayed directions.

\section{Conclusion}
In this paper, we rethink VLM adaptation by shifting from traditional weight-update paradigms to structural selection, introducing Mask Fine-Tuning (MFT). Rather than destructively updating pre-trained weights or injecting additive modules that impose capacity bottlenecks, MFT structurally selects and reorganizes the redundant representational pathways within the frozen backbone. 
Our theory establishes MFT's controlled statistical complexity and gives explicit sufficient conditions for a strictly tighter generalization bound than dense FFT. Empirically, MFT consistently outperforms FFT and PEFT across architectures, scales, and benchmarks while achieving lower training cost and zero inference overhead. Together, these results establish structural selection as an effective and general VLM adaptation paradigm.

\newpage










\bibliography{iclr2027_conference}
\bibliographystyle{iclr2027_conference}

\newpage

\appendix
\section*{Appendix}

\begin{table*}[htbp]
    \centering
    \small
    \caption{Overview of datasets used across the pretraining, SFT, and evaluation stages. Training set sizes are denoted by the number of images, while evaluation sets are denoted by the number of query instances (QAs).}
    \label{tab:dataset_splits}

    \begin{tabular*}{\textwidth}{@{\extracolsep{\fill}}lcc@{}}
        \toprule
        \textbf{Dataset Name} & \textbf{Samples} & \textbf{Usage} \\
        \midrule

        \multicolumn{3}{c}{\textit{Pretraining}} \\
        \midrule
        LAION-CC-SBU-558K~\citep{liu2023llava}
        & 558,128 & Training \\

        \midrule
        \multicolumn{3}{c}{\textit{Supervised Fine-Tuning (SFT)}} \\
        \midrule
        LAION-CC-SBU-558K~\citep{liu2023llava}
        & 558,128 & Training \\
        COCO (train2017)~\citep{lin2015microsoftcococommonobjects}
        & 118,287 & Training \\
        GQA~\citep{ainslie2023gqatraininggeneralizedmultiquery}
        & 113,018 & Training \\
        OCR-VQA~\citep{mishraICDAR19}
        & 207,572 & Training \\
        TextVQA~\citep{singh2019towards}
        & 28,408 & Training \\
        VisualGenome~\citep{krishna2016visualgenomeconnectinglanguage}
        & 108,077 & Training \\

        \midrule
        \multicolumn{3}{c}{\textit{Evaluation}} \\
        \midrule
        GQA~\citep{ainslie2023gqatraininggeneralizedmultiquery}
        & 12,578 & Evaluation \\
        MMMU~\citep{yue2023mmmu}
        & 11,500 & Evaluation \\
        POPE~\citep{Li-hallucination-2023}
        & 3,000 & Evaluation \\
        MME~\citep{fu2025mmecomprehensiveevaluationbenchmark}
        & 2,374 & Evaluation \\
        SQA~\citep{lu2022learn}
        & 4,241 & Evaluation \\
        TextVQA~\citep{singh2019towards}
        & 5,000 & Evaluation \\

        \bottomrule
    \end{tabular*}
\end{table*}

\section{Implementation Details}
\label{sec:implementation}
\subsection{Dataset Details}
The training and evaluation datasets in our experiments are identical to those used by LLaVA~\citep{liu2023llava} and TinyLLaVA~\citep{zhou2024tinyllava}. Tab.~\ref{tab:dataset_splits} summarizes the dataset splits.

\subsection{Training Details}
Our experiments are conducted on a server equipped with 4 NVIDIA A100 GPUs with 80GB of memory.
We borrow the code of LLaVA~\citep{liu2023llava} and TinyLLaVA~\citep{zhou2024tinyllava}, as well as their LLaVA-v1.5-Mix-665K instruction dataset for projector pre-training and the BLIP-Laion-CC-SBU-558K instruction dataset for fine-tuning.
We also use the official evaluation protocols for both codebases in all our experiments.

\section{Experiments}
\subsection{Complete Main Results}
We present our main results in Tab.~\ref{tab:exps_supp}, including additional fine-tuning methods (LoRA-One~\citep{zhang2025loraoneonestepgradientsuffice}, Uni-LoRA~\citep{li2025uniloravectorneed}, Adapter~\citep{houlsby2019parameter}, MoReS~\citep{bi-etal-2025-llava}, IA³~\citep{liu2022fewshotparameterefficientfinetuningbetter}). The results demonstrate that our S-MFT consistently outperforms all other methods, aligning with our claim in the main text.
\begin{table*}[!htbp]
    \caption{
    Main experimental result comparison with the time cost of our Soft Mask Fine-Tuning (S-MFT) with Full Fine-Tuning (FFT) and representative Parameter-Efficient Fine-Tuning (PEFT) baselines across four language backbones. We repeat each experiment five times with different random seeds and report the mean and standard deviation. \textcolor[RGB]{235,85,75}{Red numbers} denote the best results, and \textcolor[RGB]{75,115,210}{blue numbers} denote the second-best. All results reported here are the performance after a full training epoch. ``Epochs'' and ``Time'' are reported as the efficiency metric, indicating when a method converges and achieves stable, optimal validation performance.
    }
    \centering
    \setlength{\tabcolsep}{8pt}
    \adjustbox{width=\textwidth}{
    \begin{tabular}{llccclcccc} 
        \toprule
        \multicolumn{2}{c}{\multirow{2}{*}{\textbf{Method}}} 
        & \multicolumn{6}{c}{\textbf{Evaluation Tasks}} 
        & \multirow{2}{*}{\textbf{Epoch} {↓}}
        & \multirow{2}{*}{\textbf{Time (h)} {↓}} \\ 
        \cmidrule(lr){3-8}
        \multicolumn{2}{l}{} 
        & \multicolumn{1}{c}{\textbf{GQA} {↑}} 
        & \multicolumn{1}{c}{\textbf{MMMU} {↑}} 
        & \multicolumn{1}{c}{\textbf{POPE} {↑}} 
        & \multicolumn{1}{c}{\textbf{MME} {↑}} 
        & \multicolumn{1}{c}{\textbf{SQA-Image} {↑}} 
        & \multicolumn{1}{c}{\textbf{TextVQA} {↑}}
        & 
        & \\ 
        \midrule
        \multirow{15}{*}{\rotatebox[origin=c]{90}{\textbf{Qwen2.5-0.5B}}}
        & Zero-Shot
        & 5.7 
        & 22.6 
        & 49.9 
        & \multicolumn{1}{c}{93.7} 
        & 3.2 
        & 9.0 
        & - 
        & - \\
        & FFT
        & 58.3$_{\pm0.28}$ 
        & 33.3$_{\pm0.37}$ 
        & 86.1$_{\pm0.46}$ 
        & 1253.0$_{\pm10.2}$ 
        & 58.9$_{\pm0.25}$ 
        & 48.2$_{\pm0.23}$ 
        & 0.90 
        & 4.91 \\
        & LoRA
        & 55.6$_{\pm0.28}$ 
        & 32.0$_{\pm0.35}$ 
        & 86.9$_{\pm0.45}$ 
        & 1206.0$_{\pm10.1}$ 
        & 58.6$_{\pm0.22}$ 
        & 45.5$_{\pm0.24}$ 
        & 0.65 
        & 3.14 \\
        & DoRA
        & 57.1$_{\pm0.24}$ 
        & 32.8$_{\pm0.31}$ 
        & \textcolor[RGB]{75,115,210}{\textbf{87.1}}$_{\pm0.42}$ 
        & 1228.5$_{\pm9.4}$ 
        & 58.8$_{\pm0.21}$ 
        & 46.8$_{\pm0.28}$ 
        & 0.63 
        & 3.42 \\
        & QLoRA
        & 53.8$_{\pm0.36}$ 
        & 30.1$_{\pm0.41}$ 
        & 85.2$_{\pm0.47}$ 
        & 1178.4$_{\pm10.2}$ 
        & 56.9$_{\pm0.27}$ 
        & 46.7$_{\pm0.39}$ 
        & 0.88 
        & 5.97 \\
        & LoRA-One
        & 56.0$_{\pm0.25}$ 
        & 29.9$_{\pm0.32}$ 
        & 87.0$_{\pm0.49}$ 
        & 1208.4$_{\pm9.5}$ 
        & 57.2$_{\pm0.20}$ 
        & 45.5$_{\pm0.23}$ 
        & 0.88 
        & 4.14 \\
        & Uni-LoRA
        & 52.5$_{\pm0.21}$ 
        & 32.0$_{\pm0.34}$ 
        & 86.0$_{\pm0.46}$ 
        & 1227.3$_{\pm10.9}$ 
        & 58.0$_{\pm0.29}$ 
        & 42.4$_{\pm0.39}$ 
        & 0.69 
        & 3.98 \\
        & Adapter
        & 56.4$_{\pm0.26}$ 
        & 31.3$_{\pm0.35}$ 
        & 82.5$_{\pm0.44}$ 
        & 1284.8$_{\pm10.5}$ 
        & 56.2$_{\pm0.23}$ 
        & 44.4$_{\pm0.31}$ 
        & 0.75
        & 4.02 \\
        & Parallel Adapter
        & 56.0$_{\pm0.25}$ 
        & 30.8$_{\pm0.32}$ 
        & 86.5$_{\pm0.49}$ 
        & 1215.4$_{\pm9.5}$ 
        & 57.6$_{\pm0.20}$ 
        & 45.8$_{\pm0.23}$ 
        & 0.72 
        & 3.95 \\
        & P-Tuning v2
        & 52.8$_{\pm0.21}$ 
        & 28.5$_{\pm0.34}$ 
        & 84.1$_{\pm0.46}$ 
        & 1162.3$_{\pm10.9}$ 
        & 54.2$_{\pm0.29}$ 
        & 42.4$_{\pm0.39}$ 
        & 0.85 
        & 3.58 \\
        & MoReS 
        & 57.7$_{\pm0.24}$ 
        & 32.1$_{\pm0.33}$ 
        & 84.8$_{\pm0.41}$ 
        & \textcolor[RGB]{75,115,210}{\textbf{1310.5}}$_{\pm9.2}$ 
        & 57.8$_{\pm0.26}$ 
        & 46.3$_{\pm0.29}$ 
        & 0.78
        & 4.15 \\
        & IA³
        & 55.3$_{\pm0.31}$ 
        & 30.5$_{\pm0.42}$ 
        & 81.3$_{\pm0.48}$ 
        & 1259.8$_{\pm11.2}$ 
        & 55.1$_{\pm0.28}$ 
        & 43.2$_{\pm0.35}$ 
        & 0.68
        & 3.15 \\
        & \textbf{S-MFT Attn(Ours)} 
        & 58.0$_{\pm0.27}$ 
        & 33.3$_{\pm0.31}$ 
        & 86.7$_{\pm0.47}$ 
        & 1243.0$_{\pm8.2}$ 
        & \textcolor[RGB]{75,115,210}{\textbf{59.7}}$_{\pm0.20}$ 
        & 47.9$_{\pm0.37}$ 
        & \textcolor[RGB]{235,85,75}{\textbf{0.61}} 
        & \textcolor[RGB]{235,85,75}{\textbf{2.92}}\\
        & \textbf{S-MFT MLP (Ours)} 
        & \textcolor[RGB]{75,115,210}{\textbf{59.0}}$_{\pm0.21}$ 
        & \textcolor[RGB]{75,115,210}{\textbf{33.4}}$_{\pm0.32}$ 
        & \textcolor[RGB]{235,85,75}{\textbf{87.3}}$_{\pm0.45}$ 
        & 1306.0$_{\pm8.4}$ 
        & \textcolor[RGB]{235,85,75}{\textbf{60.0}}$_{\pm0.17}$ 
        & \textcolor[RGB]{75,115,210}{\textbf{48.9}}$_{\pm0.39}$ 
        & \textcolor[RGB]{75,115,210}{\textbf{0.62}} 
        & \textcolor[RGB]{75,115,210}{\textbf{3.13}} \\
        & \textbf{S-MFT Both(Ours)} 
        & \textcolor[RGB]{235,85,75}{\textbf{59.5}}$_{\pm0.22}$ 
        & \textcolor[RGB]{235,85,75}{\textbf{34.8}}$_{\pm0.34}$ 
        & 87.0$_{\pm0.48}$ 
        & \textcolor[RGB]{235,85,75}{\textbf{1349.3}}$_{\pm8.6}$ 
        & 59.2$_{\pm0.19}$ 
        & \textcolor[RGB]{235,85,75}{\textbf{49.4}}$_{\pm0.38}$ 
        & 0.67 
        & 3.81 \\
        \midrule
        \multirow{15}{*}{\rotatebox[origin=c]{90}{\textbf{TinyLlama-1.1B}}}
        & Zero-Shot
        & 1.0 
        & 26.1 
        & 49.8 
        & \multicolumn{1}{c}{211.8} 
        & 4.0 
        & 6.8 
        & - 
        & - \\
        & FFT
        & 57.3$_{\pm0.33}$ 
        & 30.4$_{\pm0.34}$ 
        & 85.8$_{\pm0.44}$ 
        & 1180.9$_{\pm9.2}$ 
        & 60.3$_{\pm0.27}$ 
        & 47.9$_{\pm0.23}$ 
        & 0.93 
        & 5.76 \\
        & LoRA
        & 57.5$_{\pm0.41}$ 
        & 30.4$_{\pm0.22}$ 
        & 86.9$_{\pm0.47}$ 
        & 1167.6$_{\pm9.5}$ 
        & 58.2$_{\pm0.19}$ 
        & 47.9$_{\pm0.27}$ 
        & 0.77 
        & 4.53 \\
        & DoRA
        & 57.4$_{\pm0.31}$ 
        & \textcolor[RGB]{75,115,210}{\textbf{30.9}}$_{\pm0.25}$ 
        & 87.1$_{\pm0.43}$ 
        & 1175.2$_{\pm9.8}$ 
        & 59.8$_{\pm0.18}$ 
        & 48.1$_{\pm0.26}$ 
        & 0.72 
        & 4.81 \\
        & QLoRA
        & 55.9$_{\pm0.35}$ 
        & 26.2$_{\pm0.35}$ 
        & 81.5$_{\pm0.48}$ 
        & 1142.8$_{\pm11.5}$ 
        & 56.5$_{\pm0.22}$ 
        & 46.2$_{\pm0.38}$ 
        & 0.62 
        & 5.03 \\
        & LoRA-One
        & 57.2$_{\pm0.30}$ 
        & 29.4$_{\pm0.31}$ 
        & 87.2$_{\pm0.48}$ 
        & 1170.5$_{\pm8.9}$ 
        & 59.5$_{\pm0.21}$ 
        & 46.2$_{\pm0.24}$ 
        & 0.66 
        & 4.15\\
        & Uni-LoRA
        & 55.4$_{\pm0.27}$ 
        & 27.1$_{\pm0.28}$ 
        & 84.5$_{\pm0.44}$ 
        & 1122.0$_{\pm11.8}$ 
        & 54.7$_{\pm0.14}$ 
        & 44.4$_{\pm0.22}$ 
        & 0.63 
        & 3.92 \\
        & Adapter
        & 56.6$_{\pm0.32}$ 
        & 29.5$_{\pm0.29}$ 
        & 86.2$_{\pm0.45}$ 
        & 1163.4$_{\pm9.6}$ 
        & 57.3$_{\pm0.25}$ 
        & 46.5$_{\pm0.27}$ 
        & 0.84 
        & 4.68 \\
        & Parallel Adapter
        & 56.8$_{\pm0.30}$ 
        & 29.8$_{\pm0.31}$ 
        & 86.4$_{\pm0.48}$ 
        & 1165.5$_{\pm8.9}$ 
        & 57.5$_{\pm0.21}$ 
        & 46.8$_{\pm0.24}$ 
        & 0.81 
        & 4.62 \\
        & P-Tuning v2
        & 53.4$_{\pm0.27}$ 
        & 25.1$_{\pm0.28}$ 
        & 82.5$_{\pm0.44}$ 
        & 1105.0$_{\pm11.8}$ 
        & 52.7$_{\pm0.14}$ 
        & 44.4$_{\pm0.22}$ 
        & 0.88 
        & 4.25 \\
        & MoReS 
        & 57.7$_{\pm0.38}$ 
        & 30.6$_{\pm0.24}$ 
        & 87.3$_{\pm0.44}$ 
        & 1185.6$_{\pm8.8}$ 
        & 59.2$_{\pm0.17}$ 
        & 48.0$_{\pm0.25}$ 
        & 0.75 
        & 4.85 \\
        & IA³
        & 54.2$_{\pm0.31}$ 
        & 26.5$_{\pm0.35}$ 
        & 83.1$_{\pm0.51}$ 
        & 1118.2$_{\pm10.2}$ 
        & 53.8$_{\pm0.19}$ 
        & 45.1$_{\pm0.32}$ 
        & 0.68 
        & 3.75 \\
        & \textbf{S-MFT Attn(Ours)} 
        & 57.1$_{\pm0.20}$ 
        & 30.7$_{\pm0.29}$ 
        & 87.6$_{\pm0.47}$ 
        & 1170.9$_{\pm7.4}$ 
        & 60.4$_{\pm0.15}$ 
        & \textcolor[RGB]{235,85,75}{\textbf{49.1}}$_{\pm0.37}$ 
        & \textcolor[RGB]{75,115,210}{\textbf{0.61}} 
        & \textcolor[RGB]{235,85,75}{\textbf{2.98}} \\
        & \textbf{S-MFT MLP (Ours)} 
        & \textcolor[RGB]{75,115,210}{\textbf{57.9}}$_{\pm0.30}$ 
        & \textcolor[RGB]{75,115,210}{\textbf{30.9}}$_{\pm0.20}$ 
        & \textcolor[RGB]{235,85,75}{\textbf{88.0}}$_{\pm0.48}$ 
        & \textcolor[RGB]{235,85,75}{\textbf{1237.9}}$_{\pm8.2}$ 
        & \textcolor[RGB]{235,85,75}{\textbf{62.6}}$_{\pm0.16}$ 
        & 47.4$_{\pm0.35}$ 
        & \textcolor[RGB]{235,85,75}{\textbf{0.60}} 
        & \textcolor[RGB]{75,115,210}{\textbf{3.65}} \\
        & \textbf{S-MFT Both(Ours)} 
        & \textcolor[RGB]{235,85,75}{\textbf{58.1}}$_{\pm0.22}$ 
        & \textcolor[RGB]{235,85,75}{\textbf{32.0}}$_{\pm0.32}$ 
        & \textcolor[RGB]{75,115,210}{\textbf{87.7}}$_{\pm0.47}$ 
        & \textcolor[RGB]{75,115,210}{\textbf{1218.0}}$_{\pm7.7}$ 
        & \textcolor[RGB]{75,115,210}{\textbf{61.0}}$_{\pm0.19}$ 
        & \textcolor[RGB]{75,115,210}{\textbf{48.7}}$_{\pm0.39}$ 
        & 0.74 
        & 4.32 \\
        \midrule
        \multirow{15}{*}{\rotatebox[origin=c]{90}{\textbf{Gemma-2B}}}
        & Zero-Shot
        & 2.6 
        & 26.4 
        & 56.6 
        & \multicolumn{1}{c}{586.2} 
        & 2.8 
        & 5.6 
        & - 
        & - \\
        & FFT
        & 60.9$_{\pm0.37}$ 
        & \textcolor[RGB]{75,115,210}{\textbf{31.3}}$_{\pm0.35}$ 
        & 86.4$_{\pm0.44}$ 
        & 1339.0$_{\pm10.8}$ 
        & 60.5$_{\pm0.27}$ 
        & 50.7$_{\pm0.26}$ 
        & 0.88 
        & 13.98 \\
        & LoRA
        & 58.3$_{\pm0.32}$ 
        & 29.6$_{\pm0.30}$ 
        & 86.2$_{\pm0.46}$ 
        & 1178.0$_{\pm9.7}$ 
        & 57.4$_{\pm0.20}$ 
        & 48.0$_{\pm0.23}$ 
        & 0.64 
        & 9.43 \\
        & DoRA
        & 59.6$_{\pm0.28}$ 
        & 30.8$_{\pm0.32}$ 
        & 86.8$_{\pm0.41}$ 
        & 1250.4$_{\pm9.5}$ 
        & 59.1$_{\pm0.22}$ 
        & 49.8$_{\pm0.24}$ 
        & 0.61 
        & 9.85 \\
        & QLoRA
        & 56.7$_{\pm0.36}$ 
        & 29.9$_{\pm0.39}$ 
        & 85.1$_{\pm0.46}$ 
        & 1152.3$_{\pm10.4}$ 
        & 55.8$_{\pm0.25}$ 
        & 46.1$_{\pm0.33}$ 
        & 0.63 
        & 10.71 \\
        & LoRA-One
        & 58.4$_{\pm0.32}$ 
        & 30.1$_{\pm0.32}$ 
        & 86.0$_{\pm0.46}$ 
        & 1224.7$_{\pm10.4}$ 
        & 58.4$_{\pm0.22}$ 
        & 49.5$_{\pm0.25}$ 
        & 0.70 
        & 8.19 \\
        & Uni-LoRA
        & 56.6$_{\pm0.29}$ 
        & 30.4$_{\pm0.32}$ 
        & 83.5$_{\pm0.43}$ 
        & 1194.7$_{\pm10.9}$ 
        & 57.4$_{\pm0.20}$ 
        & 49.9$_{\pm0.25}$ 
        & 0.71 
        & 9.10 \\
        & Adapter
        & 58.6$_{\pm0.33}$ 
        & 30.3$_{\pm0.35}$ 
        & 86.1$_{\pm0.48}$ 
        & 1215.3$_{\pm10.1}$ 
        & 57.8$_{\pm0.25}$ 
        & 48.6$_{\pm0.28}$ 
        & 0.75 
        & 9.92 \\
        & Parallel Adapter
        & 58.8$_{\pm0.32}$ 
        & 30.5$_{\pm0.32}$ 
        & 86.4$_{\pm0.46}$ 
        & 1218.7$_{\pm10.4}$ 
        & 58.0$_{\pm0.22}$ 
        & 48.8$_{\pm0.25}$ 
        & 0.72 
        & 9.75 \\
        & P-Tuning v2
        & 54.6$_{\pm0.29}$ 
        & 27.8$_{\pm0.32}$ 
        & 82.5$_{\pm0.43}$ 
        & 1114.7$_{\pm10.9}$ 
        & 53.4$_{\pm0.20}$ 
        & 45.9$_{\pm0.25}$ 
        & 0.82 
        & 8.50 \\
        & MoReS 
        & 59.8$_{\pm0.29}$ 
        & 31.0$_{\pm0.31}$ 
        & 87.0$_{\pm0.42}$ 
        & 1265.5$_{\pm9.3}$ 
        & 59.6$_{\pm0.21}$ 
        & 50.1$_{\pm0.26}$ 
        & 0.65 
        & 10.15 \\
        & IA³
        & 55.4$_{\pm0.35}$ 
        & 28.5$_{\pm0.38}$ 
        & 83.6$_{\pm0.47}$ 
        & 1135.8$_{\pm11.2}$ 
        & 54.2$_{\pm0.24}$ 
        & 46.5$_{\pm0.31}$ 
        & 0.71 
        & 8.15 \\
        & \textbf{S-MFT Attn (Ours)} 
        & 60.5$_{\pm0.24}$ 
        & 31.2$_{\pm0.30}$ 
        & \textcolor[RGB]{75,115,210}{\textbf{87.4}}$_{\pm0.45}$ 
        & 1348.8$_{\pm8.6}$ 
        & \textcolor[RGB]{235,85,75}{\textbf{63.5}}$_{\pm0.24}$ 
        & 50.3$_{\pm0.21}$ 
        & \textcolor[RGB]{235,85,75}{\textbf{0.53}} 
        & \textcolor[RGB]{235,85,75}{\textbf{7.21}} \\
        & \textbf{S-MFT MLP (Ours)} 
        & \textcolor[RGB]{75,115,210}{\textbf{61.4}}$_{\pm0.26}$ 
        & \textcolor[RGB]{235,85,75}{\textbf{31.9}}$_{\pm0.31}$ 
        & \textcolor[RGB]{75,115,210}{\textbf{87.4}}$_{\pm0.47}$ 
        & \textcolor[RGB]{235,85,75}{\textbf{1402.6}}$_{\pm9.8}$ 
        & \textcolor[RGB]{75,115,210}{\textbf{62.4}}$_{\pm0.22}$ 
        & \textcolor[RGB]{75,115,210}{\textbf{51.1}}$_{\pm0.21}$ 
        & \textcolor[RGB]{75,115,210}{\textbf{0.57}} 
        & \textcolor[RGB]{75,115,210}{\textbf{8.06}} \\
        & \textbf{S-MFT Both (Ours)} 
        & \textcolor[RGB]{235,85,75}{\textbf{61.6}}$_{\pm0.25}$ 
        & \textcolor[RGB]{235,85,75}{\textbf{31.9}}$_{\pm0.28}$ 
        & \textcolor[RGB]{235,85,75}{\textbf{87.6}}$_{\pm0.49}$ 
        & \textcolor[RGB]{75,115,210}{\textbf{1361.8}}$_{\pm9.1}$ 
        & 61.2$_{\pm0.20}$ 
        & \textcolor[RGB]{235,85,75}{\textbf{52.3}}$_{\pm0.22}$ 
        & 0.58 
        & 8.16 \\
        \midrule
        \multirow{15}{*}{\rotatebox[origin=c]{90}{\textbf{Phi-2-2.7B}}}
        & Zero-Shot
        & 2.3 
        & 25.7 
        & 50.1 
        & \multicolumn{1}{c}{87.4} 
        & 9.8 
        & 1.5 
        & - 
        & - \\
        & FFT
        & 60.5$_{\pm0.37}$ 
        & 38.2$_{\pm0.32}$ 
        & 86.4$_{\pm0.45}$ 
        & 1413.9$_{\pm9.6}$ 
        & \textcolor[RGB]{75,115,210}{\textbf{71.9}}$_{\pm0.29}$ 
        & 56.5$_{\pm0.22}$ 
        & 0.86 
        & 13.53 \\
        & LoRA
        & 59.8$_{\pm0.34}$ 
        & 37.1$_{\pm0.22}$ 
        & 88.1$_{\pm0.49}$ 
        & 1382.3$_{\pm9.7}$ 
        & 71.1$_{\pm0.20}$ 
        & 53.7$_{\pm0.25}$ 
        & 0.67 
        & 8.15 \\
        & DoRA
        & 61.0$_{\pm0.31}$ 
        & 37.9$_{\pm0.26}$ 
        & 88.2$_{\pm0.41}$ 
        & 1395.2$_{\pm8.8}$ 
        & 71.5$_{\pm0.23}$ 
        & 55.2$_{\pm0.27}$ 
        & 0.65 
        & 8.65 \\
        & QLoRA
        & 58.3$_{\pm0.39}$ 
        & 33.9$_{\pm0.28}$ 
        & 85.7$_{\pm0.47}$ 
        & 1354.6$_{\pm10.3}$ 
        & 69.4$_{\pm0.28}$ 
        & 56.9$_{\pm0.37}$ 
        & 0.72 
        & 11.18 \\
        & LoRA-One
        & 60.1$_{\pm0.34}$ 
        & 35.9$_{\pm0.20}$ 
        & 85.4$_{\pm0.45}$ 
        & 1380.7$_{\pm9.7}$ 
        & 69.6$_{\pm0.27}$ 
        & 53.0$_{\pm0.24}$ 
        & 0.60 
        & 7.69 \\
        & Uni-LoRA
        & 59.0$_{\pm0.33}$ 
        & 35.0$_{\pm0.39}$ 
        & 86.5$_{\pm0.47}$ 
        & 1370.6$_{\pm9.6}$ 
        & 70.3$_{\pm0.28}$ 
        & \textcolor[RGB]{75,115,210}{\textbf{58.0}}$_{\pm0.31}$ 
        & 0.72 
        & 9.94 \\
        & Adapter
        & 60.1$_{\pm0.36}$ 
        & 36.5$_{\pm0.22}$ 
        & 86.6$_{\pm0.48}$ 
        & 1385.2$_{\pm10.1}$ 
        & 70.5$_{\pm0.29}$ 
        & 54.1$_{\pm0.26}$ 
        & 0.71 
        & 8.55 \\
        & Parallel Adapter
        & 60.3$_{\pm0.34}$ 
        & 36.8$_{\pm0.20}$ 
        & 86.8$_{\pm0.45}$ 
        & 1388.7$_{\pm9.7}$ 
        & 70.8$_{\pm0.27}$ 
        & 54.4$_{\pm0.24}$ 
        & 0.69 
        & 8.42 \\
        & P-Tuning v2
        & 57.0$_{\pm0.33}$ 
        & 31.5$_{\pm0.39}$ 
        & 84.5$_{\pm0.47}$ 
        & 1280.6$_{\pm9.6}$ 
        & 66.3$_{\pm0.28}$ 
        & 51.0$_{\pm0.31}$ 
        & 0.81 
        & 7.84 \\
        & MoReS 
        & 61.2$_{\pm0.30}$ 
        & 38.0$_{\pm0.24}$ 
        & 88.2$_{\pm0.43}$ 
        & 1410.5$_{\pm9.2}$ 
        & 71.6$_{\pm0.22}$ 
        & 55.8$_{\pm0.29}$ 
        & 0.66 
        & 8.85 \\
        & IA³
        & 57.8$_{\pm0.37}$ 
        & 32.5$_{\pm0.41}$ 
        & 85.1$_{\pm0.50}$ 
        & 1305.4$_{\pm10.8}$ 
        & 67.5$_{\pm0.32}$ 
        & 51.8$_{\pm0.35}$ 
        & 0.75 
        & 7.35 \\
        & \textbf{S-MFT Attn(Ours)} 
        & 60.3$_{\pm0.24}$ 
        & 37.8$_{\pm0.27}$ 
        & \textcolor[RGB]{75,115,210}{\textbf{88.3}}$_{\pm0.45}$ 
        & 1403.9$_{\pm9.3}$ 
        & \textcolor[RGB]{235,85,75}{\textbf{72.4}}$_{\pm0.20}$ 
        & 56.1$_{\pm0.25}$ 
        & \textcolor[RGB]{235,85,75}{\textbf{0.46}} 
        & \textcolor[RGB]{235,85,75}{\textbf{5.21}} \\
        & \textbf{S-MFT MLP (Ours)} 
        & \textcolor[RGB]{75,115,210}{\textbf{61.4}}$_{\pm0.25}$ 
        & \textcolor[RGB]{75,115,210}{\textbf{38.6}}$_{\pm0.25}$ 
        & 88.0$_{\pm0.47}$ 
        & \textcolor[RGB]{235,85,75}{\textbf{1468.3}}$_{\pm9.1}$ 
        & 71.7$_{\pm0.23}$ 
        & \textcolor[RGB]{235,85,75}{\textbf{58.8}}$_{\pm0.31}$ 
        & \textcolor[RGB]{75,115,210}{\textbf{0.54}} 
        & \textcolor[RGB]{75,115,210}{\textbf{7.33}} \\
        & \textbf{S-MFT Both(Ours)} 
        & \textcolor[RGB]{235,85,75}{\textbf{61.9}}$_{\pm0.26}$ 
        & \textcolor[RGB]{235,85,75}{\textbf{39.1}}$_{\pm0.28}$ 
        & \textcolor[RGB]{235,85,75}{\textbf{88.8}}$_{\pm0.45}$ 
        & \textcolor[RGB]{75,115,210}{\textbf{1444.5}}$_{\pm9.0}$ 
        & \textcolor[RGB]{235,85,75}{\textbf{72.4}}$_{\pm0.21}$ 
        & 57.2$_{\pm0.28}$ 
        & 0.64 
        & 8.12 \\
        \bottomrule
    \end{tabular}
    }
    \label{tab:exps_supp}
\end{table*}

\subsection{Learning Rate Ablation}
\label{sec:lr}
S-MFT initializes every mask entry close to one, so the initial effective weights are close to the pre-trained weights:
\begin{equation}
W\odot M_{\mathrm{init}}
\approx W\odot\mathbf 1=W,
\end{equation}
where $\mathbf 1$ denotes the all-ones matrix. The exact initial mask is $\sigma(S_{\mathrm{init}}/T)$ and is strictly below one for finite $S_{\mathrm{init}}$. Because the score gradient is multiplied coordinate-wise by $W\odot\sigma_T'(S)$ in Eq.~\ref{eq:gradient-sensitivity}, initialization and temperature directly affect the scale of score updates. We therefore report the learning-rate ablation in Fig.~\ref{fig:lr-ablation}; hyperparameter selection must use a held-out validation split that is disjoint from every reported evaluation set.

\subsection{Data Efficiency.} 
\begin{figure*}[t]
    \centering
    \includegraphics[width=1\linewidth]{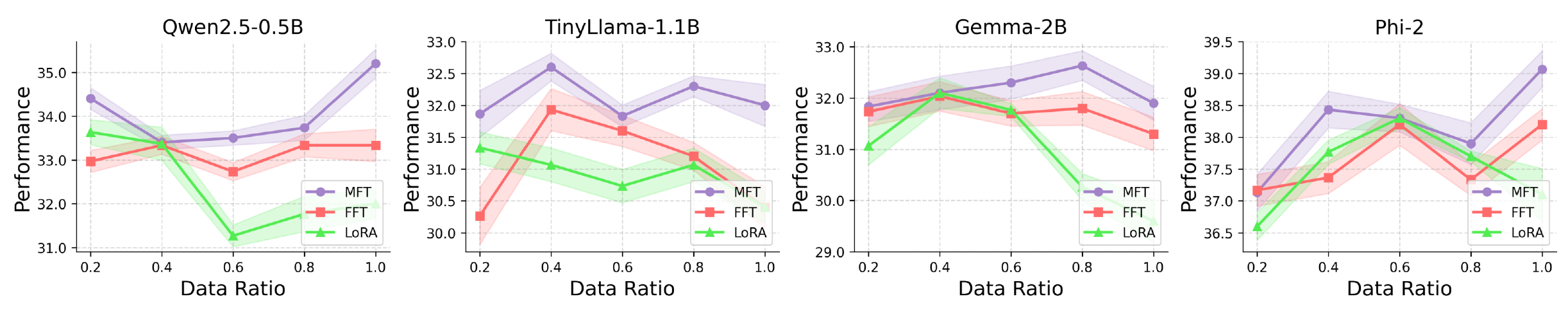}
    \caption{
    Performance comparison of S-MFT, FFT, and LoRA under varying training data ratios across four language backbones. Each curve shows the MMMU benchmark performance, with shaded regions indicating the standard deviation. S-MFT consistently outperforms FFT and LoRA across varying data ratios, with particularly pronounced advantages in low-data regimes.
    }
    \label{fig:data-ratio}
\end{figure*}

We further evaluate S-MFT's data efficiency by varying the fraction of training data used during fine-tuning (Fig.~\ref{fig:data-ratio}). Across all four backbones, S-MFT consistently outperforms FFT and LoRA under different data ratios. The advantage becomes particularly pronounced in low-data regimes, suggesting that the structural selection learned by S-MFT enables more efficient adaptation when supervision is limited.

\section{Analysis}

\subsection{Mathematical Proofs}
\label{sec:appendix_theory}

This appendix proves the claims in Sec.~\ref{sec:theory}. The probability statements are conditional on the fixed pre-trained weights $W_0$ and assume i.i.d.\ downstream examples and a per-example loss in $[0,1]$. All mask and FFT quantization alphabets and all candidate compression configurations used by the compression bounds are fixed independently of the downstream training sample.

\noindent\textbf{Technical setup and full statements.}
Let $\mathcal S=\{z_i\}_{i=1}^{n}$ be drawn i.i.d.\ from
$\mathcal D$, and define
\begin{equation}
\scalebox{0.9}{$
R(h)=\mathbb E_{z\sim\mathcal D}[\ell(h,z)],
\qquad
\widehat R_{\mathcal S}(h)
=\frac{1}{n}\sum_{i=1}^{n}\ell(h,z_i)
$},
\end{equation}
where $\ell(h,z)\in[0,1]$ is a bounded evaluation loss or a
pre-specified clipped and rescaled training loss. The pre-trained
weights $W_0$ are treated as fixed before $\mathcal S$ is observed.
The matched comparison in Sec.~\ref{sec:theory} uses the same $d$
mask-eligible coordinates and the same effective update budget for MFT
and FFT, isolating the statistical benefit of structural selection.

For a radius $r$ fixed independently of $\mathcal S$, applying the
standard empirical Rademacher bound to the two classes in
Eq.~\ref{eq:class-inclusion} and taking a union bound gives, with
probability at least $1-\delta$, simultaneously for
$A\in\{\mathrm{MFT},\mathrm{FFT}\}$ and all
$h\in\mathcal H_A(r)$,
\begin{equation}
R(h)
\le
\widehat R_{\mathcal S}(h)
+2\widehat{\mathfrak R}_n
\!\left(\ell\circ\mathcal H_A(r)\right)
+3\sqrt{\frac{\ln(4/\delta)}{2n}}.
\label{eq:rad-generalization}
\end{equation}
Together with Eq.~\ref{eq:rad-inclusion}, this establishes the
no-larger complexity term claimed in the main text. If $r$ is chosen
after observing $\mathcal S$ from a finite grid $\mathcal R$ fixed in
advance, the same argument replaces $\ln(4/\delta)$ by
$\ln(4|\mathcal R|/\delta)$.

\noindent\textbf{Compression codebook and special cases.}
For fixed integers $s\le d$ and $b\ge1$, let
$\mathcal C_{s,b}$ contain masks that select $s$ coordinate slots,
set every unselected value to one, and assign each selected coordinate
one of at most $2^b$ values from a quantization alphabet fixed before
observing $\mathcal S$. Then
\begin{equation}
|\mathcal C_{s,b}|
\le \binom{d}{s}2^{bs}.
\label{eq:mask-code-cardinality}
\end{equation}
The bound in Eq.~\ref{eq:soft-compression-bound} is stated for fixed
$s$, $b$, and alphabet. If $(s,b)$ is selected using $\mathcal S$ from
a finite candidate set $\mathcal K$ declared in advance, the numerator
acquires an additional $\ln|\mathcal K|$ term. If the loss is
$G$-Lipschitz in the effective weights on the relevant parameter set,
the distortion in the main text satisfies
\begin{equation}
\alpha(M,\widetilde M)
\le
G\|W_0\odot(M-\widetilde M)\|_2.
\label{eq:compression-distortion}
\end{equation}

For H-MFT with exactly $s$ zeros and all other mask entries equal to
one, value quantization is unnecessary and $\alpha=0$. Thus,
\begin{equation}
\scalebox{0.9}{$
R(h_{\mathrm{H\text{-}MFT}})
\le
\widehat R_{\mathcal S}(h_{\mathrm{H\text{-}MFT}})
+\sqrt{
\frac{\ln \binom{d}{s}+\ln(1/\delta)}{2n}}
$}.
\label{eq:hard-compression-bound}
\end{equation}
The combinatorial term already accounts for selecting the mask from
the training sample. Selecting $s$ from a predeclared finite candidate
set requires the same additional $\ln|\mathcal K|$ term.

For comparison, fix a dense $b_F$-bit, coordinate-wise code for FFT,
with at most $2^{b_Fd}$ codewords. For a coded approximation
$\widetilde W_F$ to the learned FFT weights, let
$\alpha_F=\sup_z|\ell(h_F,z)-\ell(h_{\widetilde W_F},z)|$. The same
finite-class argument gives
\begin{equation}
R(h_F)
\le
\widehat R_{\mathcal S}(h_F)
+2\alpha_F
+\sqrt{
\frac{b_Fd\ln2+\ln(1/\delta)}
{2n}}.
\label{eq:fft-compression-bound}
\end{equation}
If $b_F$ or its alphabet is selected using $\mathcal S$, the
corresponding candidate-set term is included. The strict comparison
evaluates the complete right-hand sides of
Eqs.~\ref{eq:soft-compression-bound} and
\ref{eq:fft-compression-bound} using the bounded empirical losses,
compression distortions, and stated code lengths. This protocol
directly quantifies when MFT's shorter structural code translates into
a tighter generalization guarantee.

\noindent\textbf{Proof of Lemma.} \textit{The MFT reparameterization function $\Phi(S) = W_0 \odot \sigma_T(S)$ is globally Lipschitz continuous with constant $L_\Phi = \frac{\|W_0\|_\infty}{4T}$.}

\textit{Proof.} Let $S \in \mathbb{R}^d$. The Jacobian matrix of the transformation $\Phi(S)$ is a diagonal matrix $J_\Phi(S) \in \mathbb{R}^{d \times d}$, where the $i$-th diagonal element is given by:
\begin{equation}
    \scalebox{0.9}{$
    [J_\Phi(S)]_{ii} = \frac{\partial \Phi_i}{\partial S_i} = W_{0,i} \cdot \frac{1}{T} \sigma_T(S_i)\bigl( 1 - \sigma_T(S_i) \bigr)
    $}.
\end{equation}
The derivative of the standard sigmoid function $\sigma(z)$ attains its global maximum of $1/4$ at $z=0$. Therefore, for the temperature-scaled sigmoid, we have $\max_{S_i} \frac{\partial \sigma_T(S_i)}{\partial S_i} = \frac{1}{4T}$.

\begin{figure*}[t]
    \centering
    
    \begin{minipage}[t]{0.48\linewidth}
        \centering
        \includegraphics[width=\linewidth]{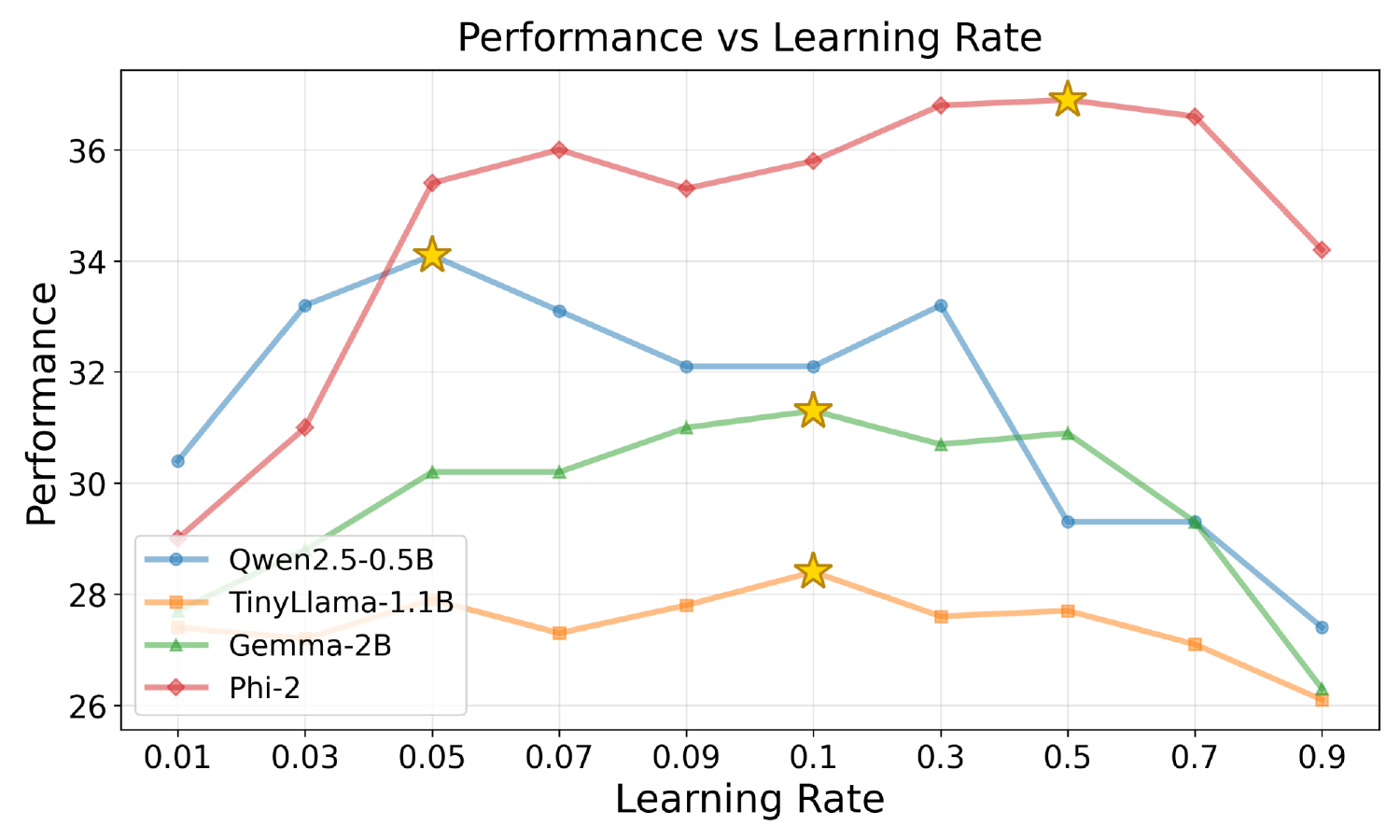}
        \caption{
        Performance across four language backbones on the MMMU with different learning rates.
        }
        \label{fig:lr-ablation}
    \end{minipage}
    \hfill
    \begin{minipage}[t]{0.48\linewidth}
        \centering
        \raisebox{0.5mm}{\includegraphics[width=\linewidth]{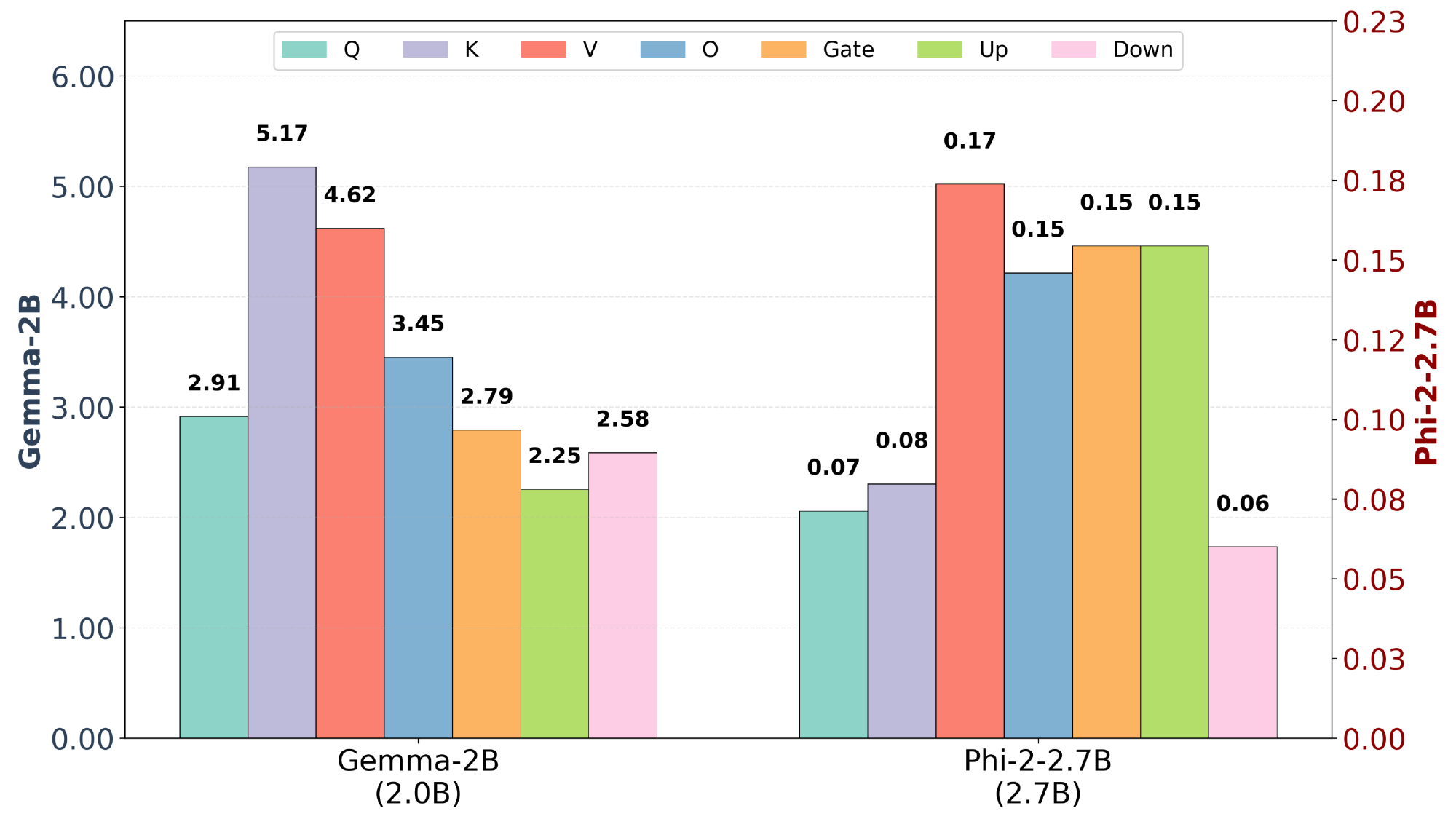}}
        \caption{
        Learned mask distributions for two language backbones, Gemma-2B and Phi-2-2.7B.
        }
        \label{fig:mask-near0-supp}
    \end{minipage}
\end{figure*}

To find the Lipschitz constant $L_\Phi$, we bound the spectral norm (maximum singular value) of the Jacobian matrix:
\begin{equation}
    \scalebox{0.8}{$
    L_\Phi = \sup_{S} \|J_\Phi(S)\|_2 = \sup_{S} \max_i \left| W_{0,i} \cdot \frac{1}{4T} \right| = \frac{\|W_0\|_\infty}{4T}
    $}.
\end{equation}
For any $S_1,S_2\in\mathbb R^d$, define the line segment
$\gamma(t)=S_2+t(S_1-S_2)$ for $t\in[0,1]$. The fundamental theorem of
calculus applied along this path gives
\begin{equation}
\Phi(S_1)-\Phi(S_2)
=\int_0^1J_\Phi(\gamma(t))(S_1-S_2)\,dt.
\end{equation}
Taking Euclidean norms and using the uniform Jacobian bound yields
\begin{equation}
\scalebox{0.9}{$
\begin{aligned}
\|\Phi(S_1)-\Phi(S_2)\|_2
&\le
\int_0^1\|J_\Phi(\gamma(t))\|_2\,dt\,
\|S_1-S_2\|_2\\
&\le
\frac{\|W_0\|_\infty}{4T}\|S_1-S_2\|_2
\end{aligned}
$}.
\end{equation}
which proves Eq.~\ref{eq:revised-lipschitz}.

\noindent\textbf{Gradient-sensitivity consequence.}
For any differentiable empirical objective $\mathcal L(W')$, the multivariate chain rule and the diagonal form of $J_\Phi(S)$ give
\begin{equation}
\scalebox{0.9}{$
\nabla_S\mathcal L
=J_\Phi(S)^\top\nabla_{W'}\mathcal L
=\left(\nabla_{W'}\mathcal L\odot W_0\right)
\odot\sigma_T'(S)
$}.
\end{equation}
Submultiplicativity of the Euclidean operator norm and the preceding lemma yield
\begin{equation}
\scalebox{0.9}{$
\|\nabla_S\mathcal L\|_2
\le\|J_\Phi(S)\|_2
\|\nabla_{W'}\mathcal L\|_2
\le\frac{\|W_0\|_\infty}{4T}
\|\nabla_{W'}\mathcal L\|_2
$}.
\end{equation}
This proves Eq.~\ref{eq:gradient-sensitivity} and explicitly bounds the score-space sensitivity contributed by the MFT reparameterization relative to the effective-weight gradient.

\noindent\textbf{Proof of the update-class inclusion.}
For any $M\in[0,1]^d$, write
\begin{equation}
\Delta W_M=W_0\odot(M-\mathbf1).
\end{equation}
Coordinate-wise, $M_i-1\in[-1,0]$, and hence
$\Delta W_{M,i}W_{0,i}=W_{0,i}^2(M_i-1)\le0$
and
$|\Delta W_{M,i}|\le|W_{0,i}|$.
If additionally $\|\Delta W_M\|_2\le r$, then $\Delta W_M$ belongs to the Euclidean ball
$\mathcal U_{\mathrm{FFT}}(r)$ by definition. Therefore
$\mathcal U_{\mathrm{MFT}}(r)\subseteq
\mathcal U_{\mathrm{FFT}}(r)$.
Applying the same map
$\Delta W\mapsto[x\mapsto f(x;W_0+\Delta W)]$
to both sets proves
$\mathcal H_{\mathrm{MFT}}(r)\subseteq
\mathcal H_{\mathrm{FFT}}(r)$.

For a fixed sample, empirical Rademacher complexity is
\begin{equation}
\widehat{\mathfrak R}_n(\mathcal G)
=
\mathbb E_{\xi}
\left[
\sup_{g\in\mathcal G}
\frac1n\sum_{i=1}^{n}\xi_i g(z_i)
\right],
\end{equation}
where the $\xi_i$ are independent Rademacher variables. Since a supremum over a subset cannot exceed the supremum over its superset, applying this definition to the two loss classes proves Eq.~\ref{eq:rad-inclusion}.
For each fixed loss class, the standard empirical Rademacher theorem for $[0,1]$-valued functions gives the corresponding inequality in Eq.~\ref{eq:rad-generalization} with failure probability $\delta/2$. A union bound over the two classes replaces $\ln(2/(\delta/2))$ by $\ln(4/\delta)$ and proves that Eq.~\ref{eq:rad-generalization} holds simultaneously for MFT and FFT.
If the radius is selected from a predeclared finite grid
$\mathcal R$, allocating failure probability
$\delta/(2|\mathcal R|)$ to each method--radius pair gives the stated
$\ln(4|\mathcal R|/\delta)$ post-selection term.

\noindent\textbf{Proof of the compression bound.}
Let $\mathcal C_{s,b}$ be the fixed coded-mask class from Eq.~\ref{eq:mask-code-cardinality}, and write
$N_{s,b}=|\mathcal C_{s,b}|$.
For any fixed $\widetilde M\in\mathcal C_{s,b}$, Hoeffding's inequality for the bounded loss gives
\begin{equation}
\Pr\!\left[
R(h_{\widetilde M})
-\widehat R_{\mathcal S}(h_{\widetilde M})
>t
\right]
\le\exp(-2nt^2).
\end{equation}
Taking a union bound over all $N_{s,b}$ coded masks and setting
\begin{equation}
t=
\sqrt{\frac{\ln N_{s,b}+\ln(1/\delta)}{2n}}
\end{equation}
shows that, with probability at least $1-\delta$, simultaneously for all
$\widetilde M\in\mathcal C_{s,b}$,
\begin{equation}
R(h_{\widetilde M})
\le
\widehat R_{\mathcal S}(h_{\widetilde M})
+
\sqrt{\frac{\ln N_{s,b}+\ln(1/\delta)}{2n}}.
\label{eq:finite-code-bound-proof}
\end{equation}
This event is uniform over the codebook, so it remains valid when the training sample selects the coded mask.
If $(s,b)$ is itself selected from a predeclared finite set
$\mathcal K$, another union bound allocates failure probability
$\delta/|\mathcal K|$ to each configuration and adds
$\ln|\mathcal K|$ to the numerator. This proves the post-selection variant stated after Eq.~\ref{eq:soft-compression-bound}.

Now let $M$ be any soft mask and
$\widetilde M\in\mathcal C_{s,b}$ any coded approximation. By the definition of
$\alpha(M,\widetilde M)$,
\begin{align}
R(h_M)
&\le R(h_{\widetilde M})
+\alpha(M,\widetilde M),\\
\widehat R_{\mathcal S}(h_{\widetilde M})
&\le\widehat R_{\mathcal S}(h_M)
+\alpha(M,\widetilde M).
\end{align}
Combining these two inequalities with
Eq.~\ref{eq:finite-code-bound-proof} and using
$N_{s,b}\le\binom{d}{s}2^{bs}$
proves Eq.~\ref{eq:soft-compression-bound}.

If $\ell(f(\cdot;W),z)$ is $G$-Lipschitz in $W$ uniformly over $z$ on the line segments joining the original and coded effective weights, then
\begin{equation}
\begin{aligned}
\alpha(M,\widetilde M)
&\le
G\|
W_0\odot M-W_0\odot\widetilde M
\|_2\\
&=
G\|W_0\odot(M-\widetilde M)\|_2,
\end{aligned}
\end{equation}
which proves Eq.~\ref{eq:compression-distortion}.

For an H-MFT mask with exactly $s$ zeros, the selected coordinates and their fixed zero values completely determine the mask. There are $\binom{d}{s}$ such masks, and choosing
$\widetilde M=M$ makes $\alpha=0$. Substitution in
Eq.~\ref{eq:finite-code-bound-proof} proves
Eq.~\ref{eq:hard-compression-bound}.

Finally, a fixed dense $b_F$-bit FFT code has at most
$2^{b_Fd}$ codewords. Repeating the same uniform Hoeffding argument with its loss distortion $\alpha_F$ proves Eq.~\ref{eq:fft-compression-bound}. Comparing the complete right-hand sides gives the stated sufficient criterion: MFT's shorter structural code, together with preserved empirical fit and small approximation distortion, yields a strictly tighter bound than dense FFT.

\begin{figure}[t]
    \centering
    \includegraphics[width=1\linewidth]{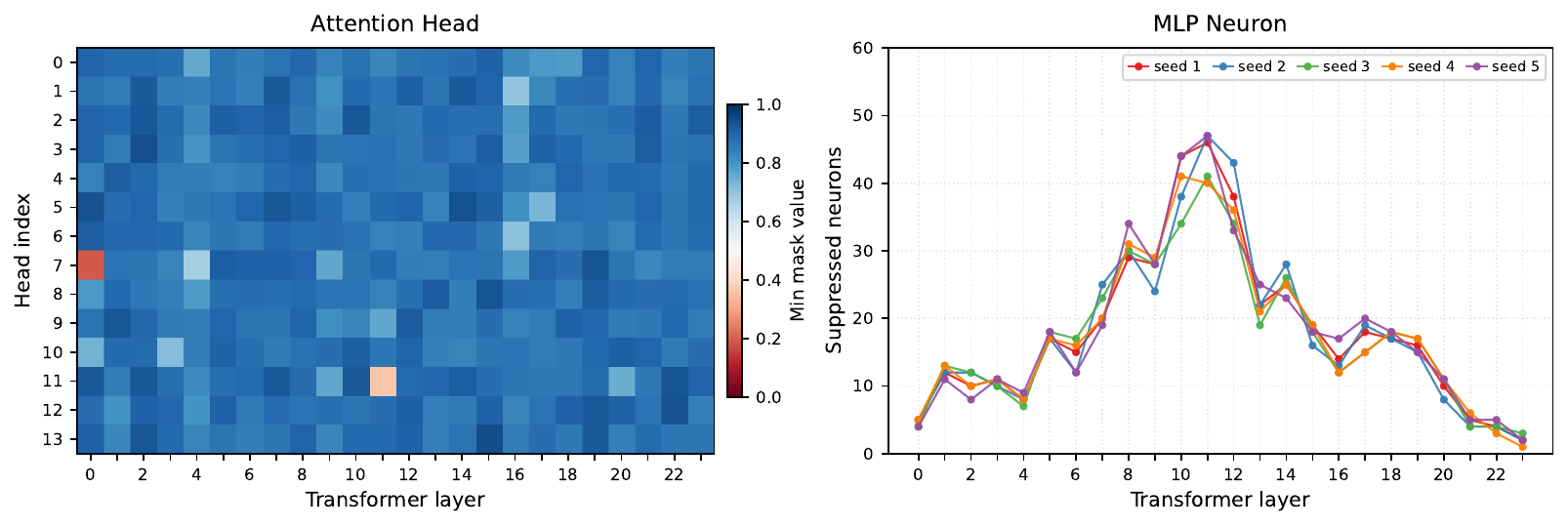}
    \vspace{-8mm}
    \caption{Analysis of block-level masks across 5 seeds.
    }
    \label{fig:block-level masks}
\end{figure}

\subsection{Additional Mask Analysis}
Fig.~\ref{fig:mask-near0-supp} shows that Gemma-2B has a larger proportion of mask values below $0.1$ than Phi-2-2.7B in the displayed projections. These architecture-specific attenuation signatures show that S-MFT adapts its structural selection to each backbone's organization. Targeted intervention studies that remove, restore, or exchange selected connections provide a promising direction for further characterizing the causal roles of these pathways.

\subsection{Block-level masks.}
We use weight-level masks, given their flexibility, to explore the effectiveness of structural selection. We add Block-level S-MFT on LLaVA-Qwen2.5-0.5B for comparison, which learns masks on Attn heads and MLP neurons.
We directly transfer the best hyperparameters from the weight-level setting to make an initial attempt, evaluated on 4 tasks.

\begin{wraptable}{r}{0.52\columnwidth}
\centering
\footnotesize
\vspace{-10pt}
\caption{Mask granularity comparison on LLaVA-Qwen2.5-0.5B.}
\vspace{-6pt}
\renewcommand{\arraystretch}{0.8}
\resizebox{\linewidth}{!}{%
\setlength{\tabcolsep}{4pt}
\begin{tabular}{lcccc}
\toprule
Method & MMMU $\uparrow$ & POPE $\uparrow$ & SQA $\uparrow$ & TextVQA $\uparrow$ \\
\midrule
Block & 31.8 & 86.5 & 58.7 & 43.7 \\
Weight & \textbf{35.1} & \textbf{87.5} & \textbf{59.4} & \textbf{49.6} \\
\bottomrule
\end{tabular}
}
\label{tab:mask_granularity}
\vspace{-10pt}
\end{wraptable}

\noindent Because block-level masks are less flexible than weight-level masks, their lower performance is expected. But the moderate gap suggests it may be improved with dedicated hyperparameter tuning. We further ran both methods with the same hyperparameters across 5 seeds. Weight-level masks are stable in overall sparsity, not in individual weight positions, even though performance is similar.
In contrast, block-level masks show stable unit-level patterns. One Attn unit (layer 0, head 7) is always strongly suppressed, while most others remain active. Redundancy is mainly observed in MLP neurons, with about 400 neurons suppressed per run, concentrated in the middle layers (especially layers 10-12, in Fig.~\ref{fig:block-level masks}). These reproducible unit-level patterns provide a more interpretable view of where redundancy is concentrated. Understanding why these specific units are selected requires a deeper mechanistic analysis, which is beyond the scope of this submission. We believe this is a meaningful research direction, as block-level masks may also be more hardware-friendly.

\subsection{Inference \& Deployment.} 
The learned masks are merged offline by multiplying them directly into the frozen weights ($W'=W\odot M$), guaranteeing zero inference overhead. H-MFT additionally supports lightweight task-specific binary masks alongside a shared backbone, substantially reducing multi-task storage compared with maintaining separate fine-tuned models or adapters.

\subsection{Expressivity through Pretrained-Structure Preservation.}
S-MFT parameterizes a targeted effective-weight family that preserves each pre-trained connection's sign while learning its task-specific contribution through continuous attenuation. This geometry produces the class inclusion in Eq.~\ref{eq:class-inclusion} and the resulting statistical-complexity advantage. Across all evaluated tasks, this pretrained, structure-preserving family consistently outperforms solutions obtained by FFT and PEFT. The architecture-specific patterns in Figs.~\ref{fig:mask-ratio} and~\ref{fig:mask-near0-supp} further show that MFT deploys its capacity selectively, adapting the location and strength of attenuation to each backbone.

\end{document}